\documentclass[a4paper,11pt]{article}
\usepackage[colorlinks=false]{hyperref}

\usepackage{url}
\usepackage{xcolor}
\definecolor{newcolor}{rgb}{.8,.349,.1}

\usepackage{amsmath, bm, amsthm, amssymb, mathrsfs}
\usepackage{amsfonts}
\usepackage{xcolor}
\usepackage{graphicx}
\usepackage{booktabs, multirow}
\usepackage{enumitem}
\usepackage{array} 
\usepackage{caption} 

\definecolor{darkgreen}{RGB}{0 155 0}

\newcommand\tr{\intercal}
\newcommand{\ve}[1]{\bm{#1}}
\newcommand{\vx}{\ve{x}}
\newcommand{\vX}{\ve{X}}
\newcommand{\vy}{\ve{y}}
\newcommand{\vY}{\ve{Y}}
\newcommand{\vz}{\ve{z}}
\newcommand{\vZ}{\ve{Z}}
\newcommand{\vu}{\ve{u}}
\newcommand{\vU}{\ve{U}}

\newcommand\xdim{d}
\newcommand\FXext{(F_{1}(x_{1}),\,F_{2}(x_{2}),\ldots,\,F_{\xdim}(x_{\xdim}))}
\newcommand\FXinvext{(F_{1}^{-1}(u_{1}),\,F_{2}^{-1}(u_{2}),\ldots,\,F_{\xdim}^{-1}(u_{\xdim}))}
\newcommand\xext{(x_{1},\,x_{2},\ldots,\,x_{\xdim})}
\newcommand\uext{(u_{1},\,u_{2},\ldots,\,u_{\xdim})}

\newcommand\dX{\bm{\mathcal{X}}}
\newcommand\dY{\mathcal Y}
\newcommand\dZ{\bm{\mathcal Z}}
\newcommand\dU{\bm{\mathcal U}}
\newcommand\dXY{(\dX, \, \dY)}
\newcommand\dXYprime{(\dX^\prime, \, \dY^\prime)}
\newcommand\dXYsecond{(\dX'', \, \dY'')}

\newcommand\MODEL{\mathcal M}
\newcommand{\PC}{\textrm{PC}}
\newcommand\MPC{\MODEL_{\PC}}
\newcommand\T{\mathcal T}

\newcommand\TU{\T^{\textrm{PIT}}}
\newcommand{\TR}{\T}

\newcommand\cm{\,\textrm{cm}}

\newcommand\A{\mathcal{A}}
\newcommand\notA{\overline{\A}}

\newcommand{\R}{\mathbb R}
\newcommand{\E}{\mathbb E}
\newcommand{\V}{\mathbb V}

\newcommand\cdf{\mathrm{CDF}}
\newcommand\pdf{\mathrm{PDF}}
\newcommand\AIC{\mathrm{AIC}}
\newcommand\TPRa{\mathcal{T}^{(\Pi)}}
\newcommand{\Nw}{\,\mathrm{N}}

\newcommand\fX{f_{\vX}}
\newcommand\FX{F_{\vX}}
\newcommand\CX{C_{\vX}}
\newcommand\cX{c_{\vX}}

\newcommand\FZ{F_{\vZ}}

\newcommand\Ran{\textrm{Ran}}

\newcommand\Psialpha{\Psi_{\bm{\alpha}}}
\newcommand\Psibeta{\Psi_{\bm{\beta}}}
\newcommand\phialpha{\phi_{\alpha_{i}}^{(i)}}
\newcommand\phibeta{\phi_{\beta_{i}}^{(i)}}

\newcommand\FcondA{F_{\notA|\A}}
\newcommand\fcondA{f_{\notA|\A}}

\newcommand\ccondA{c_{\notA|\A}}

\newcommand\MWh{\,\textrm{MWh}}
\newcommand\MAE{\textrm{MAE}}

\newcommand\rMAE{\textrm{rMAE}}

\newcommand\eqrefp[1]{(\ref{eq:#1})}
\newcommand\secref[1]{Section~\ref{sec:#1}}
\newcommand\subsecref[1]{Section~\ref{subsec:#1}}
\newcommand\subsubsecref[1]{Section~\ref{subsubsec:#1}}
\newcommand\appref[1]{\ref{app:#1}}
\newcommand\subappref[1]{\ref{subapp:#1}}

\newcommand\figref[1]{Figure~\ref{fig:#1}}
\newcommand\tabref[1]{Table~\ref{tab:#1}}

\newcommand\ie{\textit{i.e.}}
\newcommand\eg{\textit{e.g.}}

\theoremstyle{plain}
\newtheorem*{thm*}{\protect\theoremname}
\theoremstyle{plain}

\providecommand{\theoremname}{Theorem}

\newcommand\aPCEonX{\textit{aPCEonX}} 
\newcommand\lPCEonX{\textit{lPCEonX}} 
\newcommand\lPCEonZ{\textit{lPCEonZ}}

\DeclareMathOperator*{\argmin}{arg\,min}
\DeclareMathOperator*{\AP}{AP}

\title{Data-driven polynomial chaos expansion for machine learning regression}

\author{E. Torre, S. Marelli, P. Embrechts, B. Sudret}

\date{}

\usepackage{natbib}
\usepackage[left=25mm,right=25mm,top=1.5cm,bottom=1.5cm,includeheadfoot]{geometry}
\begin{document}
		
\maketitle		
		
\abstract{
	We present a regression technique for data-driven problems based on polynomial chaos expansion (PCE). PCE is a popular technique in the field of uncertainty quantification (UQ), where it is typically used to replace a runnable but expensive computational model subject to random inputs with an inexpensive-to-evaluate polynomial function. The metamodel obtained enables a reliable estimation of the statistics of the output, provided that a suitable probabilistic model of the input is available. 
	
	Machine learning (ML) regression is a research field that focuses on providing purely data-driven input-output maps, with the focus on pointwise prediction accuracy. We show that a PCE metamodel purely trained on data can yield pointwise predictions whose accuracy is comparable to that of other ML regression models, such as neural networks and support vector machines. The comparisons are performed on benchmark datasets available from the literature. The methodology also enables the quantification of the output uncertainties, and is robust to noise. Furthermore, it enjoys additional desirable properties, such as good performance for small training sets and simplicity of construction, with only little parameter tuning required.

}

	\section{Introduction} \label{sec:introduction}
	
	Machine learning (ML) is increasingly used today to make predictions of system responses and to aid or guide decision making. Given a $\xdim$-dimensional input vector $\vX$ to a system and the corresponding $q$-dimensional output vector $\vY$, data-driven ML algorithms establish a map ${\MODEL:\,\vX\mapsto\vY}$ on the basis of an available sample set ${\dX = {\vx^{(1)},\dots,\vx^{(n)}}}$ of input observations and of the corresponding output values ${\bm{\dY} = {\vy^{(1)},\dots, \vy^{(n)}}}$, where $\vy^{(i)}=\MODEL(\vx^{(i)})+\varepsilon$ and $\varepsilon$ is a noise term. In classification tasks, the output $\vY$ is discrete, that is, it can take at most a countable number of different values (the class \emph{labels}). In regression tasks, which this paper is concerned with, the output $\vY$ takes continuous values. Regression techniques include linear regression, neural networks (NN), kernel methods (such as Gaussian processes, GP), sparse kernel machines (such as support vector machines, SVM), graphical models (such as Bayesian networks and Markov random fields), and others. A comprehensive overview of these methods can be found in \cite{Bishop2006_book} and in \cite{Witten2016_DataMining}.
	
	Current research on ML algorithms focuses on various open issues. First, there is an increasing interest towards problems where the inputs to the system, and as a consequence the system's response, are uncertain (see, \eg, \cite{Chan2018_ML4UQ, Kasiviswanathan2016_inbook, Mentch2016_1}). Properly accounting for both aleatory and epistemic input uncertainties allows one to estimate the response statistics. Second, novel methods are sought to better automatize the tuning of hyperparameters, which several ML methods are highly sensitive to \cite{Snoek2012_NIPS}. Third, the paucity of data in specific problems calls for models that can be effectively trained on few observations only \cite{Forman2004_161}. Finally, complex models, while possibly yielding accurate predictions, are often difficult if not impossible to interpret.
	
	This manuscript revisits polynomial chaos expansions (PCE), a well established metamodelling technique in uncertainty quantification (UQ), as a statistical ML \cite{Vapnik:1995} regression algorithm that can deal with these challenges \cite{Sudret2015_lecture}. UQ classically deals with problems where $\vX$ is uncertain, and is therefore modelled as a random vector. As a consequence, $\vY=\MODEL(\vX)$ is also uncertain, but its statistics are unknown and have to be estimated. Differently from ML, in UQ the model $\MODEL$ is typically available (\eg, as a finite element code), and can be computed pointwise. However, $\MODEL$ is often a computationally expensive black-box model, so that a Monte Carlo approach to estimate the statistics of $\vY$ (generate a large input sample set $\dX$ to obtain a large output sample set $\bm{\dY}=\{\MODEL(\vx),\, \vx\in\dX\}$) is not feasible. PCE is an UQ spectral technique that is used in these settings to express $\vY$ as a polynomial function of $\vX$. PCE thereby allows one to replace the original computational model $\MODEL$ with an inexpensive-to-run but accurate metamodel. The metamodel can be used to derive, for instance, estimates of various statistics of $\vY$, such as its moments or its sensitivity to different components of $\vX$ \cite{Saltelli2000}, in a computationally efficient way. PCE was originally developed for independent, identically distributed input data from a Gaussian \cite{Wiener} and then a list of named distributions \cite{Xiu2002}, then generalized to marginal and joint distributions with an arbitrary functional form \cite{Soize2004, Wan2006_aPCE}. A moment-based approach that does not require a functional form of the marginals to be available (or even to exist) was proposed in \cite{Oladyshkin2012_179}, and further extended to arbitrary joint distributions in \cite{Paulson2016_3548}. In \cite{Torre2019_Vines4UQ}, we established a general framework to perform UQ (including but not limited to PCE metamodelling) in the presence of complex input dependencies modelled through copulas.
	
	Here, we re-establish PCE in a purely data-driven ML setup, where the goal is to obtain a model that can predict the response $\vY$ of a system given its inputs $\vX$. Compared to classical UQ settings where PCE is typically used, four major challenges arise in this data-driven setup:
	\begin{itemize}
		\item no computational model of the system is available, determining a significant lack of information compared to classical UQ settings;
		\item as a spectral method, PCE requires the knowledge of the joint $\pdf$ $\fX$ of the input. We address instead scenarios where $\fX$ is entirely inferred from data, a step that can affect convergence;
		\item to guarantee generality, inference on the marginal distributions is performed here non-parametrically (by kernel smoothing), the rationale being that too little evidence is available to assume any parametric families; 
		\item the obtained model needs to be robust to noise in the data, a feature that to our knowledge has not been investigated for PCE yet.
	\end{itemize}
	
	For simplicity, we consider the case where $\vY$ is a scalar random variable $Y$. The generalisation to multivariate outputs is straightforward. In the setup of interest here, no computational model $\MODEL$ is available, but only a set $\dXY$ of input values and corresponding responses. $\dX$ and $\dY$ are considered to be (possibly noisy) realisations of $\vX$ and $Y$, the true relationship between which is deterministic but unknown.
	
	After recalling the underlying theory (\secref{methods}), we first show by means of simulation that data-driven PCE delivers accurate pointwise predictions (\secref{synthetic_data}). In addition, PCE also enables a reliable estimation of the statistics of the response (such as its moments and $\pdf$), thus enabling uncertainty quantification of the predictions being made. Dependencies among the input variables are effectively modelled by copula functions, specifically by vine copulas. Vines copulas have been recently investigated for general UQ applications, including PCE metamodeling, in \cite{Torre2019_Vines4UQ}. Here their applicability to data-driven settings is investigated and demonstrated. The full approach is shown to be robust to noise in the data, a feature that has not been investigated yet for PCE, and that enables denoising. Furthermore, the methodology works well in the presence of small training sets. In \secref{applications}, we further apply PCE to real data previously analyzed by other authors with different NN and/or SVM algorithms, where it achieves a comparable performance. Importantly, the construction of the PCE metamodel does not rely on fine tuning of critical hyper-parameters. This and other desirable features of the PCE modelling framework are discussed in \secref{discussion}.

	\section{Methods} \label{sec:methods}
	
	\subsection{Measures of accuracy} \label{subsec:error_types}
	
	Before introducing PCE, which is a popular metamodelling technique in UQ, as an ML technique used to make pointwise predictions, it is important to clarify the distinction between the error types that UQ and ML typically aim at minimizing.
	
	ML algorithms are generally used to predict, for any \textit{given} input, the corresponding output of the system. Their performance is assessed in terms of the distance of the prediction from the actual system response, calculated for each input and averaged over a large number of (ideally, all) possible inputs. A common error measure in this sense, also used here, is the mean absolute error (MAE). For a regression model $\MODEL$ trained on a training data set $\dXYprime$, the MAE is typically evaluated over a validation data set $\dXYsecond$ of size $n''$ by
	\begin{equation}
	\MAE = \frac{1}{n''} \sum_{({\vx}, {y}) \in \dXYsecond} \left\vert {y}-\MODEL({\vx}) \right\vert.
	\end{equation}
	
	The associated relative error considers point by point the ratio between the absolute error and the actual response, \ie,
	\begin{equation}
	\rMAE = \frac{1}{n} \sum_{(\hat{\vx}, \hat{y}) \in \dXY} \left\vert \frac{\hat{y}-\MODEL(\hat{\vx})}{\hat{y}} \right\vert = 
	\frac{1}{n} \sum_{(\hat{\vx}, \hat{y}) \in \dXY} \left\vert 1-\frac{\MODEL(\hat{\vx})}{\hat{y}} \right\vert,
	\label{eq:rMAE}
	\end{equation}
	which is well defined if $\hat{y} \neq 0$ for all $\hat{y} \in \dY$.
	
	PCE is instead typically used to draw accurate estimates of various statistics of the output -- such as its moments, its $\pdf$, confidence intervals -- given a probabilistic model $\fX$ of an uncertain input. The relative error associated to the estimates $\hat{\mu}_Y$ of $\mu_{Y}$ and $\hat{\sigma}_Y$ of $\sigma_{Y}$ is often quantified by
	\begin{equation}
	\left\vert 1-\frac{\hat{\mu}_Y}{\mu_Y} \right\vert, \quad \left\vert 1-\frac{\hat{\sigma}_Y}{\sigma_Y} \right\vert,
	\label{eq:relerr_stats}
	\end{equation}
	provided that $\mu_Y \neq 0$ and $\sigma_Y \neq 0$. A popular measure of the error made on the full response $\pdf$ $f_Y$ is the Kullback-Leibler divergence 
	\begin{equation}
	\Delta_{\textrm{KL}}(\hat{f}_Y|f_Y) = \int_y \log \left( \frac{\hat{f}_Y(y)}{f_Y(y)} \right) f_Y(y) dy,
	\label{eq:KLdiv}
	\end{equation}
	which quantifies the average difference between the logarithms of the predicted and of the true $\pdf$s. In the simulated experiments performed in \secref{synthetic_data}, we used the error measures in \eqref{eq:relerr_stats} and \eqref{eq:KLdiv} to assess the quality of the PCE estimates $\hat{\mu}_Y$, $\hat{\sigma}_Y$, and $\hat{f}_Y$. The reference solutions $\mu$, $\sigma$, and $f_Y$ were obtained by MCS on $10^7$ samples. The real data analyzed in \secref{applications}, instead, contained too few points to draw sufficiently accurate reference values. Statistical errors for these data could not be quantified.
	
	Provided a suitable model $\fX$ of the joint $\pdf$ of the input, PCE is known to converge (in the mean-square sense) very rapidly as the size of the training set increases, compared for instance to Monte-Carlo sampling \cite{Puig2002, Todor:Schwab:2007, Ernst2012_317}. This happens because PCE, which is a spectral method, efficiently propagates the probabilistic information delivered by the input model $\fX$ to the output (as explained in more details next). Nevertheless, this does not necessarily imply a similarly rapid pointwise convergence of the error, which remains to be demonstrated. Also, the speed of mean-square convergence will generally worsen if $\fX$ is unknown and needs to be inferred from data. For a discussion of various types of convergence, see \cite{Ernst2012_317} and references therein.

	\subsection{PCE in data-driven settings} \label{subsec:PCE_representation}
	
	PCE is designed to build an inexpensive-to-evaluate analytical model $Y_{PC}=\MPC(\vX)$ mapping an input random vector $\vX$ onto an output random variable $Y$ \cite{Ghanem90, Xiu2002}. 
	PCE assumes that an unknown deterministic map $\MODEL:\R^\xdim \rightarrow \R$ exists, such that $Y=\MODEL(\vX)$. $Y$ is additionally assumed to have finite variance: $\V(Y)=\int_{\R^d} \MODEL^2(\vx) \fX(\vx) d\vx < +\infty$. Under the further assumption that each $X_i$ has finite moments of all orders \cite{Ernst2012_317}, the space of square integrable functions with respect to $\fX(\cdot)$ admits a basis $\{ \Psialpha(\cdot), \bm{\alpha} \in \mathbb{N}^\xdim \}$ of polynomials orthonormal to each other with respect to the probability measure $\fX$, \ie, such that
	\begin{equation}
	\int_{\R^\xdim}\Psialpha(\vx)\Psibeta(\vx)\fX(\vx)d\vx=\delta_{\bm{\alpha\beta}}.
	\label{eq:orthogonality}
	\end{equation}
	Here, $\delta_{\bm{\alpha\beta}}$ is the Kronecker delta symbol. The element $\alpha_i$ of the multi-index $\bm{\alpha} \in \mathbb{N}^\xdim$ indicates the degree of $\Psialpha$ in the $i$-th variable, $i=1,\ldots,\xdim$. $\Psialpha$ has a total degree given by $|\bm{\alpha}|=\sum_i \alpha_i$.
	
	Thus, $Y$ admits the spectral representation
	\begin{equation}
	Y(\vX)=\sum_{\bm{\alpha}\in\mathbb{N}^{\xdim}}y_{\bm{\alpha}}\Psialpha(\vX).
	\label{eq:PCE-def}
	\end{equation}
	
	The goal of PCE regression is to determine the coefficients $y_{\bm \alpha}$ of the expansion, truncated at some polynomial order, given an initial set $\dXY$ of observations (the \emph{training set} or \emph{experimental design}). The advantage of expressing the functional relationship between the inputs and the output by an orthonormal multivariate basis lies in its spectral decay properties. Indeed, given that finite variance models/phenomenons are considered, the orthonormality of the basis implies that the squared sum of the PCE coefficients (equal to the variance, see \subsubsecref{PCE_stats_estimation}) is finite. Therefore, the orthonormal basis guarantees a rapidly decaying spectrum of the PCE coefficients. This is turn reduces the number of parameters in the PCE representation, yielding a compressed representation. The orthonormal basis is thus paramount to reduce the number of regression coefficients and thus to avoid overfitting on finite datasets. The superiority of the orthonormal basis with respect to a non-orthogonal one in these settings is demonstrated on simulated data in \subsecref{PCEvsSALSA}.
	
	In engineering applications the model $\MODEL(\cdot)$ is often directly available (\eg, as a finite element model) but computationally expensive to evaluate: in these cases, PCE is used as a surrogate to replace the true model with an inexpensive-to-evaluate metamodel. In the standard ML settings considered here, conversely, $\MODEL(\cdot)$ is unknown and has to be modelled solely on the basis of available observations $\dXY$. The primary goal of this paper is to show that the accuracy of a PCE metamodel built purely on $\dXY$ can compete with that of state-of-the-art machine learning regression models, while requiring little parameter tuning and in some cases offering additional advantages.

	\subsection{PCE for independent inputs} \label{subsec:PCE}
	
	In this section, we assume that the input random vector $\vX$ has statistically independent components. The PCE basis is built from the tensor product of univariate orthogonal polynomials. The case of inputs with dependent components is discussed later in \subsecref{PCE_for_inputs_assumed_indep}.

	\subsubsection{Orthogonalisation for independent inputs}
	
	When $\vX$ has independent components, the $\Psialpha$ can be obtained as the tensor product of $d$ univariate polynomials,
	\begin{equation} \label{eq:tensor_product}
	\Psialpha(\vx)=\prod_{i=1}^{d}\phialpha(x_{i}),
	\end{equation}
	where each $\phialpha(x_{i})$ is the element of degree $\alpha$ of a basis of univariate polynomials orthonormal with respect to the marginals 
	$f_i$ of $\vX$, that is, such that
	\begin{equation}
	\int_{\R} \phialpha(\omega)\phibeta(\omega)f_i(\omega) d\omega =\delta_{\alpha_i\beta_i}.
	\end{equation}
	
	The problem of building a basis of mutually orthonormal multivariate polynomials with respect to $\fX$ hence becomes the problem of building $d$ bases of mutually orthonormal univariate polynomials, each with respect to a marginal $f_i$. The $d$ bases can be built independently.

	\subsubsection{Specification of the marginals and PCE construction} \label{subsubsec:marginals_inference}
	
	The construction of the polynomial basis requires a model of the marginal input distributions. Families of univariate orthonormal polynomials associated to classical continuous distributions are described in \cite{Xiu2002}. An input variable $X_i$ with a different distribution $F_i$ (continuous, strictly increasing) may be transformed into a random variable $\tilde{X}_i$ with distribution $\Phi$ belonging to one of the classical families by the transformation
	\begin{equation}
	\tilde{X}_i = \Phi^{-1}(F_i(X_i)).
	\label{eq:PIT_for_aPCE}
	\end{equation}
	This relation offers a first possible (but usually not recommended) approach to build the PCE of $Y$ for inputs $\vX$ with generic continuous marginals $F_i$: transform $\vX$ into $\tilde{\vX}$ through \eqref{eq:PIT_for_aPCE}, and then build a PCE of $Y$ in terms of $\tilde{\vX}$. The PCE has to be trained on $(\tilde{\dX}, \dY)$, where $\tilde{\dX}$ is obtained from $\dX$ using \eqref{eq:PIT_for_aPCE}.
	
	A second approach, undertaken in this study, consists in directly constructing a basis of orthonormal polynomials with respect to $F_i$, by Stiltjes or Gram-Schmidt orthogonalisation \cite{Soize2004, Wan2006_aPCE}. In practice, performing PCE on the original input $\vX$ is often preferable to building the PCE on inputs transformed via \eqref{eq:PIT_for_aPCE}. Indeed, the latter is usually a highly non-linear transformation, making the map from $\tilde{\vX}$ to $Y$ more difficult to approximate by polynomials \cite{Oladyshkin2012_179}. In the applications presented in this paper, we opt for non-parametric inference of the input marginals $f_i$ by kernel density estimation (KDE) \cite{Rosenblatt1956_832, Parzen1962_1065}. Given a set $\mathcal{X}_i=\{x_i^{(1)},\ldots,x_i^{(n)}\}$ of observations of $X_i$, the kernel density estimate $\hat{f}_i$ of $f_i$ reads
	\begin{equation}
	\hat{f}_i(x) = \frac{1}{nh}\sum_{j=1}^n k\left( \frac{x-x_i^{(j)}}{h} \right),
	\end{equation}
	where $k(\cdot)$ is the kernel function and $h$ is the appropriate kernel bandwidth that is learned from the data. Different kernels are used in practice, such as the Epanechnikov or Gaussian kernels. Here we use the latter, that is, $k(\cdot)$ is selected as the standard normal $\pdf$.
	
	A third approach for PCE construction, first introduced in \cite{Oladyshkin2012_179} under the name of arbitrary PCE (aPCE), consists in constructing a basis of polynomials orthonormal to the input moments (not to the input $\pdf$ directly). The method was first introduced  for independent input variables and later extended to correlated variables in \cite{Paulson2016_3548}. aPCE offers the advantage to be more general, as it requires only the input moments and does not need (or even assume the existence of) a functional form of the input $\pdf$. An accurate estimation of moments of higher order, however, requires a large number of input samples, thus effectively limiting its applicability in the settings considered here \cite{Oladyshkin2018_137}. For this reason, we opt instead for the second approach outlined above.
	
	After estimating the input marginals by KDE, we resort to PCE to build a basis of orthonormal polynomials. The PCE metamodel is then trained on $\dXY$.

	\subsubsection{Truncation schemes}
	
	The sum in \eqrefp{PCE-def} involves an infinite number of terms. For practical purposes, it is truncated to a finite sum. Truncation is a critical step: a wrong truncation will lead to underfitting (if too many terms are removed) or to overfitting (if too many terms are retained compared to the number of available training points). The model complexity can be controlled by various truncation schemes.
	
	The standard basis truncation \cite{Xiu2002} retains the subset of terms defined by 
	\[
	\mathcal{A}^{\xdim,p}=\{\bm{\alpha}\text{\ensuremath{\in\mathbb{N}}}^{\xdim}:\,|\bm{\alpha}|\leq p\},
	\]
	where $p\in\mathbb{N}^{+}$ is the chosen maximum polynomial degree and $|\bm{\alpha}|=\sum_{i=1}^{\xdim}\alpha_{i}$ is the total degree of $\Psialpha$. Thus, only polynomials of degree up to $p$ are considered. $\mathcal{A}^{\xdim,p}$ contains $\binom{\xdim+p}{p}$ elements. To further reduce the basis size, several additional truncation schemes have been proposed. Hyperbolic truncation \cite{Blatman2011_JCP} retains the polynomials with indices in 
	\[
	\mathcal{A}^{\xdim,p,q}=\{\bm{\alpha} \in \mathcal{A}^{\xdim,p}:\,\left\Vert \bm{\alpha}\right\Vert _{q}\leq p\},
	\]
	where $q\in(0,1]$ and $\left\Vert \cdot\right\Vert _{q}$ is the $q$-norm defined by $\left\Vert \bm{\alpha}\right\Vert _{q}=\left(\sum_{i=1}^{\xdim}\alpha_{i}^{q}\right)^{1/q}$. 
	
	A complementary strategy is to set a limit to the number of non-zero elements in $\bm{\alpha}$, that is, to the number of interactions among the components $X_{i}$ of $\vX$ in the expansion. This maximum interaction truncation scheme retains the polynomials with indices in
	\[
	\mathcal{A}^{\xdim,p,r}=\{\bm{\alpha}\in\mathcal{A}^{\xdim,p}:\ \left\Vert \bm{\alpha}\right\Vert _{0}\leq r\},
	\]
	where $\left\Vert \bm{\alpha}\right\Vert _{0}=\sum_{i=1}^{\xdim}\bm{1}_{\{\alpha_{i}>0\}}$. 
	
	In our case studies below we combined hyperbolic and maximum truncation, thus retaining the expansion's polynomials with coefficients in 
	\begin{equation}
	\mathcal{A}^{\xdim,p,q,r}=\mathcal{A}^{\xdim,p,q}\cap\mathcal{A}^{\xdim,p,r}.
	\label{eq:truncation_scheme_Mpqr}
	\end{equation}
	This choice is motivated by the sparsity of effect principle, which assumes that only few meaningful interactions influence system responses and which holds in most engineering applications.
	The three hyperparameters $(p,q,r)$ can all be automatically tuned by cross-validation within pre-assigned ranges, as illustrated in \cite{Blatman2011_JCP}. To reduce the already high computational cost due to the large number of simulations performed in the current study, we fixed $q=0.75$ and tuned $p$ and $r$ only, within pre-assigned ranges $[1, p_{\max}]$ and $[1, r_{\max}]$. $p_{\max}$ and $r_{\max}$ were selected depending on the amount of training data available. Starting from simpler models (lower $p$ and $r$), more complex models were favored if yielding a lower cross-validation error. The calibration of each parameter stopped either if no increase in performance was attained in two consecutive steps, or if the maximum allowed value was reached.
	
	\subsubsection{Calculation of the coefficients}
	
	Selected a truncation scheme and the corresponding set $\A=\{\bm{\alpha}_1,\ldots,\bm{\alpha}_{|\A|} \}$ of multi-indices, the coefficients
	$y_k$ in 
	\begin{equation}
	Y_{PC}(\vX)=\sum_{k=1}^{|\A|} y_{k}\Psi_{\bm{\alpha}_k}(\vX)
	\label{eq:PCE-truncated}
	\end{equation}
	need to be determined on the basis of a set 
	$(\dX, \, \dY) = \{({\vx}^{(j)}, \, {y}^{(j)}), \, j=1,\ldots,n\}$
	of observed data. In these settings, the vector ${\bm{y}=(y_{\bm{\alpha}_1},\ldots, y_{\bm{\alpha}_{|\A|}})}$ of expansion coefficients can be determined by regression. 
	
	When the number $|\A|$ of regressors is smaller than the number $n$ of observations, $\bm{y}$ can be determined by solving the ordinary least squares (OLS) problem
	\begin{equation} \label{eq:OLS}
	\bm{y} = 
	\argmin_{\tilde{\bm{y}}} \sum_{j=1}^n \left( {y}^{(j)} - Y_{PC}({\vx}^{(j)}) \right) = \argmin_{\tilde{\bm{y}}} \sum_{j=1}^n \left( {y}^{(j)} - \sum_{k=1}^{|\A|} y_{k} \Psi_{\bm{\alpha}_k} ({\vx}^{(j)}) \right).
	\end{equation}
	
	The solution reads
	\begin{equation} \label{eq:OLSsolution}
	\bm{y} = \left( \bm{A}^T \bm{A} \right)^{-1} \bm{A}^T 
	\begin{pmatrix}
	{y}^{(1)} \\[-6pt]
	\vdots \\[-6pt]
	{y}^{(n)}
	\end{pmatrix},
	\end{equation}
	where $A_{jk}=\Psi_{\bm{\alpha}_k}({x}^{(j)})$, $j=1,\ldots,n$, $k=1,\ldots,|\A|$.
	
	The OLS problem cannot be solved when $n<|\A|$, because in this case the matrix $\bm{A}^T \bm{A}$ in \eqref{eq:OLSsolution} is not invertible. Also, OLS tends to overfit the data when $|\A|$ is large. Simpler models with fewer coefficients can be constructed by sparse regression.
	
	Least angle regression (LAR), proposed in \cite{Efron2004}, is an algorithm that achieves sparse regression by solving the regularised regression problem
	\begin{equation} 
	\bm{y} = \argmin_{\tilde{\bm{y}}} \left\lbrace \sum_{j=1}^n \left( 
	{y}^{(j)} - \sum_{k=1}^{|\A|} y_{k} \Psi_{\bm{\alpha}_k} ({\vx}^{(j)})
	\right) + \lambda ||\bm{\tilde{y}}||_1  \right\rbrace.
	\label{eq:LAR}
	\end{equation}
	The last addendum in the expression is the regularisation term, which forces the minimisation to favour sparse solutions. One advantage of using LAR is that it does not require explicit optimization with respect to the $\lambda$ parameter, which in turn is never explicitly calculated. The use of LAR in the context of PCE was initially proposed in \cite{Blatman2011_JCP}, which the reader is referred to for more details. In the applications illustrated in Sections \ref{sec:synthetic_data} and \ref{sec:applications}, we adopt LAR to determine the PCE coefficients.

	\subsubsection{Estimation of the output statistics} \label{subsubsec:PCE_stats_estimation}
	
	Given the statistical model $\FZ$ of the input and the PCE metamodel (\ref{eq:PCE-truncated}) of the input-output map, the model response $Y_{\PC}$ is not only known pointwise, but can also be characterised statistically. For instance, the orthonormality of the polynomial basis ensures that the mean and the variance of $Y_\PC$ read, respectively,
	\begin{equation}
	\E[Y_{\PC}]=y_{\bm{0}}, \quad \V[Y_{\PC}]=\sum_{\alpha\in\A\backslash\{\bm{0}\}}y_{\bm{\alpha}}^{2}.
	\label{eq:PCE-mean-var}
	\end{equation}
	
	This property provides a useful interpretation of the expansion coefficients in terms of the first two moments of the output. Other output statistics, such as the Sobol partial sensitivity indices \cite{Sobol1993}, can be obtained from the expansion coefficients analytically \cite{SudretRESS2008b}. 
	
	Higher-order moments of the output, as well as other statistics (such as the full $\pdf$ $F_Y$), can be efficiently estimated through Monte-Carlo simulation, by sampling sufficiently many realisations of $\vX$ and evaluating the corresponding PCE responses. Polynomial evaluation is computationally cheap and can be trivially vectorised, making estimation by resampling extremely efficient.
	
	\subsection{PCE for mutually dependent inputs} \label{subsec:PCE_for_inputs_assumed_indep}
	
	If the input vector $\vX$ has statistically dependent components, its joint $\pdf$ $\fX$ is not the product of the marginals and the orthogonality property \eqref{eq:orthogonality} does not hold. As a consequence, one cannot construct multivariate orthogonal polynomials by tensor product of univariate ones, as done in \eqref{eq:tensor_product}. Constructing a basis of orthogonal polynomials in this general case is still possible, for instance by Gram-Schmidt orthonormalisation \cite{Navarro2014_arxiv}. However, the procedure becomes more and more computationally demanding as the number of inputs or the expansion order grow, as it involves the numerical evaluation of higher-dimensional integrals. For this reason, we investigate two alternative strategies.
	
	The first strategy consists in ignoring the input dependencies and in building a basis of polynomials orthonormal with respect to $\prod_i f_i(x_i)$ by arbitrary PCE. This approach is labelled from here on as \aPCEonX. While not forming a basis of the space of square integrable functions with respect to $\fX$, the considered set of polynomials may still yield a similarly good pointwise approximation. 
	
	Accurate estimates of the output statistics may be obtained \emph{a posteriori} by modelling the input dependencies through copula functions, as detailed in \appref{data_driven_input_model}. The joint $\cdf$ of a random vector $\vX$ with copula distribution $\CX$ and marginals $F_i$ (here, obtained by KDE) is given by \eqref{eq:Sklar-F-C-relation}. Resampling from such a probabilistic input model yields more accurate estimates of the output statistics than resampling from the distribution $\prod_i F_i(x_i)$, that is, than ignoring dependencies.
	
	The second possible approach, presented in more details in \appref{vines_plus_PCE}, consists in modelling directly the input dependencies by copulas, and then in transforming the input vector $\vX$ into a set of independent random variables $\vZ$ with prescribed marginals (\eg, standard uniform) through the associated Rosenblatt transform $\TPRa$ \cite{Rosenblatt52}, defined in \eqref{eq:Rosenblatt-transform-Copula-1}. The PCE metamodel is then built between the transformed input vector $\vZ$ and the output $Y$. When the input marginals are standard uniform distributions, the corresponding class of orthonormal polynomials is the Legendre family, and the resulting PCE is here indicated by \lPCEonZ. The asymptotic properties of an orthonormal PCE, such as the relations \eqref{eq:PCE-mean-var}, hold. The expression of $Y$ in terms of $\vZ$ is given by the relation \eqref{eq:YTPC}.
	
	At first, the second approach seems advantageous over the first one. It involves, in a different order, the same steps: modelling the input marginals and copula, and building the PCE metamodel. By capturing dependencies earlier on, it enables the construction of a basis of mutually orthonormal polynomials with respect to the inferred joint $\pdf$ of the (transformed) inputs. It thereby provides a model with spectral properties (except for unavoidable errors due to inference, truncation of the expansion, and estimation of the expansion parameters). The experiments run in \secref{synthetic_data}, however, show that the first approach yields typically more accurate pointwise predictions (although not always better statistical estimates). This is due to the fact that it avoids mapping the input $\vX$ into independent variables $\vZ$ via the (typically strongly non-linear) Rosenblatt transform. The shortcomings of mapping dependent inputs into independent variables were noted, in classical UQ settings, by \cite{Eldred2008_1892}, and are thus extended here to data-driven settings. For a more detailed discussion, see \subsecref{validation_conclusions}.

	\section{Validation on synthetic data} \label{sec:synthetic_data}
	
	\subsection{Validation scheme} \label{subsec:validation_framework}
	We first investigate data-driven regression by PCE on two different simulated data sets. The first data set is obtained through an analytical, highly non-linear function of three variables, subject to three uncertain inputs. The second data set is a finite element model of a horizontal truss structure, subject to $10$ uncertain inputs. In both cases, the inputs are statistically dependent, and their dependence is modelled through a canonical vine (C-vine) copula (see \appref{data_driven_input_model}). 
	
	We study the performance of the two alternative approaches described in \subsecref{PCE_for_inputs_assumed_indep}: {\aPCEonX} and {\lPCEonZ}. In both cases we model the input copula as a C-vine, fully inferred from the data. For the {\aPCEonX} the choice of the copula only plays a role in the estimation of the output statistics, while for the {\lPCEonZ} it affects also the pointwise predictions. We additionally tested the performance obtained by using the true input copula (known here because it was used to generate the synthetic data) and a Gaussian copula inferred from data. We also investigated the performance obtained by using the true marginals instead of the ones fitted by KDE. Using the true copula and marginals yielded better statistical estimates, but is of little practical interest, as it would not be known in real applications. The Gaussian copula yielded generally similar or slightly worse accuracy. For brevity, we do not show these results.
	
	To assess the pointwise accuracy, we generate a training set $\dXYprime$ of increasing size $n^{\prime}$, build the PCE metamodels both by {\aPCEonX} and by {\lPCEonZ}, and evaluate their rMAE on a validation set $\dXYsecond$ of fixed size $n''=10{,}000$ points. Both $\dX'$ and $\dX''$ are sampled from the probabilistic input model with true marginals and vine copula. The statistical accuracy is quantified instead in terms of error on the mean, standard deviation, and full $\pdf$ (by the Kullback-Leibler divergence) of the true models, as defined in \eqref{eq:rMAE}-\eqref{eq:KLdiv}. Reference solutions are obtained by MC sampling from the true input model, while the PCE estimates are obtained by resampling from the inferred marginals and copula. The sample size is $10^7$ in both cases. The statistical estimates obtained by the two PCE approaches are furthermore compared to the corresponding sample estimates obtained from the same training data (sample mean, sample standard deviation, and KDE of the $\pdf$).
	
	For each value of $n^{\prime}$ and each error type, error bands are obtained as the minimum to maximum error obtained across $10$ different realisations of $\dXYprime$ and $\dXYsecond$. The minimum error is often taken in machine learning studies as an indication of the best performance that can be delivered by a given algorithm in a given task. The maximum error represents analogously the worst-case scenario.
	
	Finally, we assess the robustness to noise by adding a random perturbation $\varepsilon$ to each output observations in $\dY'$ used to train the PCE model. The noise is drawn independently for each observation from a univariate Gaussian distribution with mean $\mu_\varepsilon=0$ and prescribed standard deviation $\sigma_\varepsilon$.
	
	\subsection{Ishigami function}
	
	The first model we consider is
	\begin{equation} 
	Y = 1+\frac{\textrm{ish}(X_1, X_2, X_3) + 1+\pi^4/10}{9+\pi^4/5},
	\label{eq:Ishigami_rescaled}
	\end{equation}
	where 
	\begin{equation} \label{eq:ishigami}
	\textrm{ish}(x_1, x_2, x_3) = \sin(x_1) + 7 \sin^2(x_2) + 0.1 x_3^4 \sin(x_1)
	\end{equation}
	is the Ishigami function \cite{Ishigami90}, defined for inputs $x_i \in [-\pi, \pi]$ and taking values in $[-1-\frac{\pi^4}{10}, 8+\frac{\pi^4}{10}]$. The Ishigami function is often used as a test case in global sensitivity analysis due to its strong non-linearity and non-monotonicity. Model \eqref{eq:Ishigami_rescaled} is rescaled here to take values in the interval $[1,2]$. Rescaling does not affect the approximation accuracy of the PCE, but enables a meaningful evaluation of the rMAE \eqref{eq:rMAE} by avoiding values of $Y$ close to $0$.
	
	We model the input $\vX$ as a random vector with marginals $X_i \sim \mathcal U([-\pi, \pi])$, $i=1,2,3$, and C-vine copula with density
	\[
	\cX(u_1, u_2, u_3) = c_{12}^{(\mathcal G)}(u_1,u_2; 2)\cdot c_{13}^{(t)}(u_1,u_3; 0.5, 3).
	\]
	Here, $c_{12}^{(\mathcal G)}(\cdot,\cdot; \theta)$ and $c_{13}^{(t)}(\cdot,\cdot; \theta)$ are the densities of the pairwise Gumbel and t- copula families defined in rows$~11$ and $19$ of \tabref{pair_copula_cdfs}. Thus, $X_1$ correlates positively with both $X_2$ and $X_3$. $X_2$ and $X_3$ are also positively correlated, but conditionally independent given $X_1$. The joint $\cdf$ of $\vX$ can be obtained from its marginals and copula through \eqref{eq:Sklar-F-C-relation}. The $\pdf$ of $Y$ in response to input $\vX$, its mean $\mu_Y$ and its standard deviation $\sigma_Y$, obtained on $10^7$ sample points, are shown in the left panel of \figref{Ytruepdf_4models}.
	
	{
		\linespread{1}
		\begin{figure}[!ht]
			\begin{centering}
				\includegraphics[width=16cm]{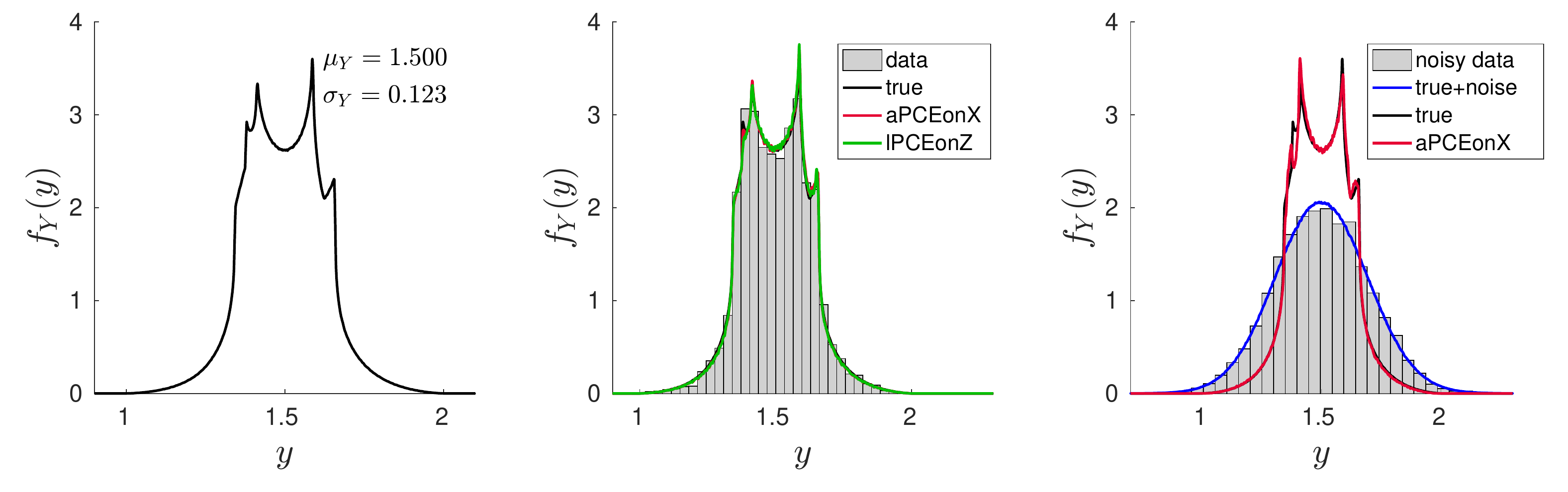}
				\caption{\textbf{Response PDFs of the Ishigami function.} \emph{Left panel:} true $\pdf$ $f_{Y}$ of the Ishigami model's response, obtained on $10^7$ sample points by KDE. \emph{Central panel:} histogram obtained from $n'=1{,}000$ output observations used for training (gray bars), true response $\pdf$ as in the left panel (black), $\pdf$s obtained from the {\aPCEonX} (red) and the {\lPCEonZ} (green) by resampling. \emph{Right panel:} as in the central panel, but for training data perturbed with Gaussian noise ($\sigma_\varepsilon=0.15=1.22\,\sigma_Y$). The blue line indicates the true $\pdf$ of the perturbed model.}
				\label{fig:Ytruepdf_4models}
			\end{centering}
		\end{figure}
		
		\begin{figure}[!ht]
			\begin{center}
				\includegraphics[width=16cm]{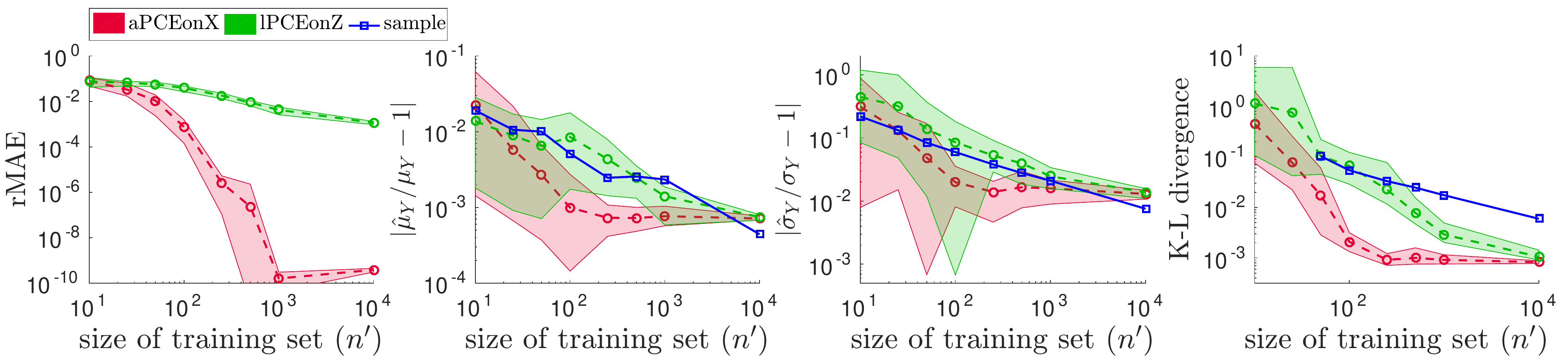}
			\end{center}
			\caption{\textbf{PCE of Ishigami model: performance}. From left to right: rMAE, error on the mean, error on the standard deviation, and Kullback-Leibler divergence of {\aPCEonX} (red) and {\lPCEonZ} (green), for a size $n'$ of the training set increasing from $10$ to $10{,}000$. The dash-dotted lines and the bands indicate, respectively, the average and the minimum to maximum errors over $10$ simulations. In the second to fourth panels, the blue lines correspond to the empirical estimates obtained from the training data (error bands not shown).
			}
			\label{fig:PCEperformance_synthetic_models}
		\end{figure}
	}
	
	{ 
		\linespread{1}
		\begin{figure}[!ht]
			\begin{center}
				\includegraphics[width=16cm]{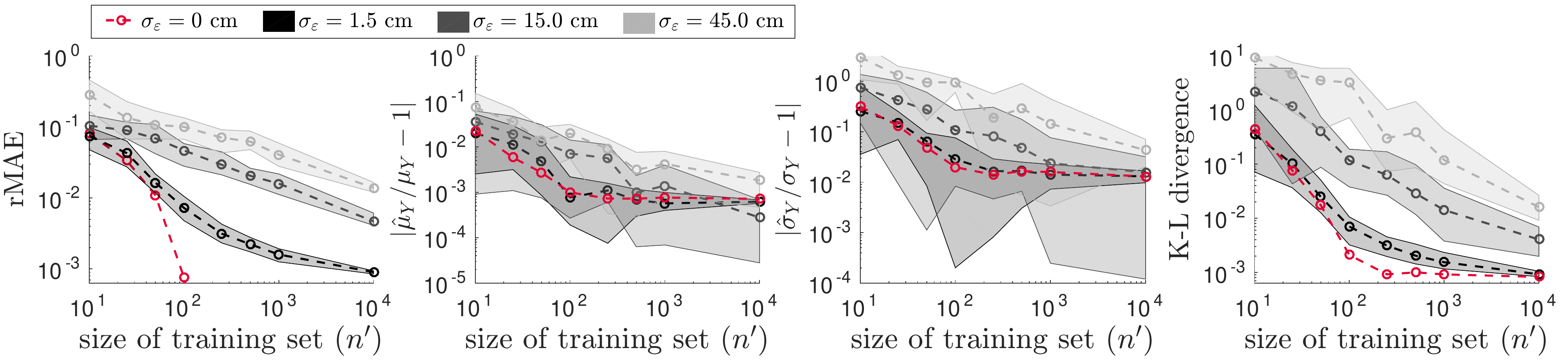}
			\end{center}
			\caption{\textbf{{\aPCEonX} of Ishigami model: robustness to noise (for multiple noise levels)}. From left to right: rMAE, error on the mean, error on the standard deviation, and Kullback-Leibler divergence of the {\aPCEonX} for an increasing amount of noise: $\sigma_\varepsilon=0.015$ (dark gray), $\sigma_\varepsilon=0.15$ (mid-light gray), and $\sigma_\varepsilon=0.45$ (light gray). The dash-dotted lines and the bands indicate, respectively, the average and the minimum to maximum error over $10$ simulations. The red lines, reported from \figref{PCEperformance_synthetic_models} for reference, indicate the mean error obtained for the noise-free data.}
			\label{fig:PCEperformance_noise_synthetic_models}
		\end{figure}
		
		\begin{figure}[!ht]
			\begin{center}
				\includegraphics[width=16cm]{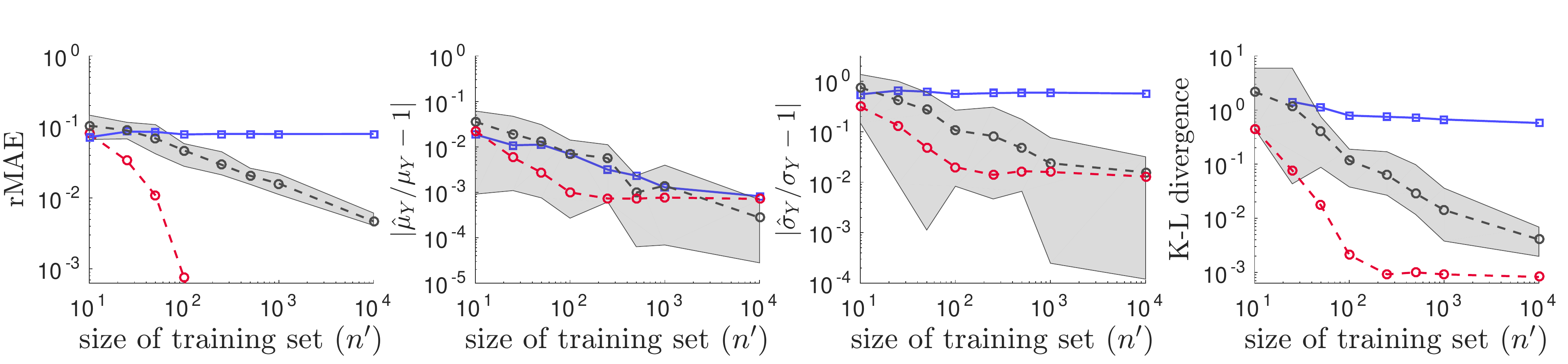}
			\end{center}
			\caption{\textbf{{\aPCEonX} of Ishigami model: robustness to noise (versus sample estimation)}. From left to right: rMAE, error on the mean, error on the standard deviation, and Kullback-Leibler divergence obtained by {\aPCEonX} (gray) and by direct sample estimation (blue), for noise ${\sigma_\varepsilon=0.15=1.22\,\sigma_Y}$. The dash-dotted lines and the bands indicate, respectively, the average and the minimum to maximum error over $10$ simulations. The red lines, reported from \figref{PCEperformance_synthetic_models} for reference, indicate the mean error obtained for the noise-free data.}
			\label{fig:PCEperformance_noise_synthetic_models_PCEvsMC}
		\end{figure}
	}
	
	Next, we build the {\aPCEonX} and the {\lPCEonZ} on training data $\dXYprime$, and assess their performance as described in \subsecref{validation_framework}. The errors are shown in \figref{PCEperformance_synthetic_models} (red: {\aPCEonX}; green: {\lPCEonZ}), as a function of the training set size $n'$. The dotted line indicates the average error over the $10$ repetitions, while the shaded area around it spans the range from the minimum to the maximum error across $10$ repetitions. The {\aPCEonX} yields a considerably lower rMAE. This is due to the strong non-linearity of the Rosenblatt transform used by the {\lPCEonZ} to de-couple the components of the input data. Importantly, the methodology works well already in the presence of relatively few data points: the pointwise error and the Kullback-Leibler divergence both drop below $1\%$ when using as few as $n'=100$ data points. The central panel of \figref{Ytruepdf_4models} shows the histogram obtained from $n'=1{,}000$ output observations of one training set, the true $\pdf$ (black), and the $\pdf$s obtained  by resampling from the {\aPCEonX} and the {\lPCEonZ} built on that training set. The statistics of the true response are better approximated by the {\aPCEonX} than by the {\lPCEonZ} or by sample estimation (blue solid lines in \figref{PCEperformance_synthetic_models}). 
	
	
	Finally, we examine the robustness of {\aPCEonX} to noise. We perturb each observation in $\dY'$ by adding noise drawn from a Gaussian distribution with mean $0$ and standard deviation $\sigma_\varepsilon$. $\sigma_\varepsilon$ is assigned as a fixed proportion of the model's true mean $\mu_Y$: $1\%$, $10\%$, and $30\%$ of $\mu_Y$ (corresponding to $12\%$, $122\%$, and $367\%$ of $\sigma_Y$, respectively). The results, shown in Figures \ref{fig:PCEperformance_noise_synthetic_models}-\ref{fig:PCEperformance_noise_synthetic_models_PCEvsMC}, indicate that the methodology is robust to noise. Indeed, the errors of all types are significantly smaller than the magnitude of the added noise, and decrease with increasing sample size (see \figref{PCEperformance_noise_synthetic_models}). For instance, the rMAE for $\sigma_\varepsilon=0.15=1.22\sigma_Y$ drops to $10^{-2}$ if $100$ or more training points are used. The error on $\mu_Y$ is minorly affected, which is expected since the noise is unbiased. More importantly, $\sigma_Y$ and $f_Y$ can be recovered with high precision even in the presence of strong noise (see also \figref{Ytruepdf_4models}, fourth panel). In this case, the PCE predictor for the standard deviation works significantly better than the sample estimates, as illustrated in \figref{PCEperformance_noise_synthetic_models_PCEvsMC}.
	

	\subsection{23-bar horizontal truss} \label{subsec:truss}
	
	We further replicate the analysis carried out in the previous section on a more complex finite element model of a horizontal truss \cite{Blatman2011_JCP}. The structure consists of $23$ bars connected at $6$ upper nodes, and is $24$ meters long and $2$ meters high (see  \figref{Truss_model_Blatman}). The bars belong to two different groups (horizontal and diagonal bars), both having uncertain Young modulus $E_i$ and uncertain cross-sectional area $A_i$, $i=1,2$:
	\begin{align*}
	E_1, E_2 \sim & \mathcal{LN} \left( 2.1 \cdot 10^{11}, 2.1 \cdot 10^{10} \right)~\textrm{Pa}, \\
	A_1 \sim & \mathcal{LN} \left( 2.0 \cdot 10^{-3} , 2.0 \cdot 10^{-4} \right)~m^2, \\
	A_2 \sim & \mathcal{LN} \left( 1.0 \cdot 10^{-3} , 1.0 \cdot 10^{-4} \right)~m^2,
	\end{align*}
	where $\mathcal{LN}(\mu, \sigma)$ is the univariate lognormal distribution with mean $\mu$ and standard deviation $\sigma$.
	
	{
		\linespread{1}
		\begin{figure}[!ht]
			\begin{center}
				\includegraphics[width=10cm]{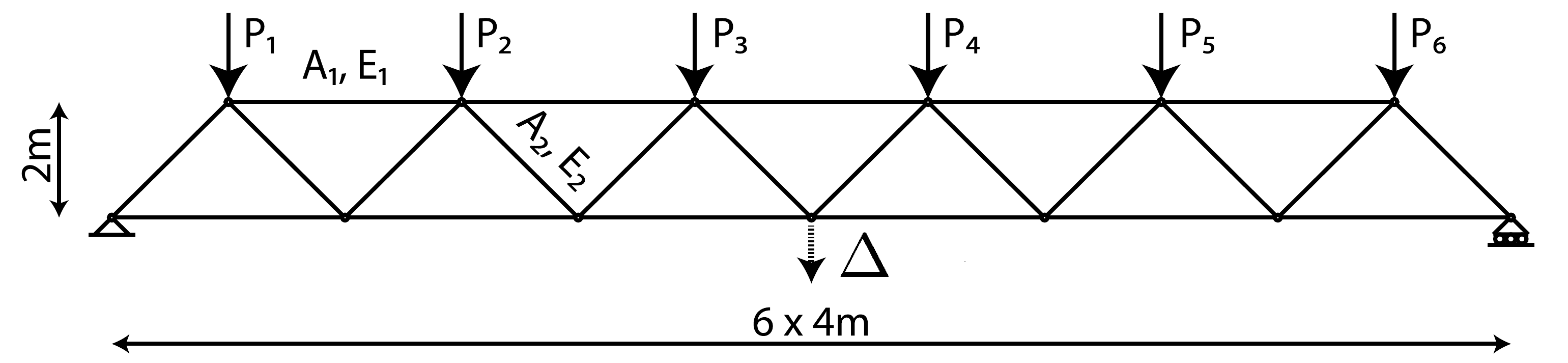}
				\caption{\textbf{Scheme of the horizontal truss model.} $23$-bar horizontal truss with bar cross-section $A_i$ and Young modulus $E_i$ ($i=1,2$: horizontal and vertical bars, respectively), subject to loads $P_1,\ldots,P_6$.  Modified from \cite{Blatman2011_JCP}.}
				\label{fig:Truss_model_Blatman}	
			\end{center}
		\end{figure}
	}
	
	The four variables can be considered statistically independent, and their values influence the structural response to loading. An additional source of uncertainty comes from six random loads $P_{1},\,P_{2},\,\ldots,\,P_{6}$ the truss is subject to, one on each upper node. The loads have Gumbel marginal distribution with mean $\mu=5\times10^{4}\Nw$ and standard deviation $\sigma=0.15\mu=7.5\times10^{3}\Nw$:
	\begin{equation}
	F_{i}(x;\alpha,\beta)=e^{-e^{-(x-\alpha)/\beta}},\quad x \in \R, \,i=1,\,2,\ldots,6,
	\label{eq:Gumbel-marginals}
	\end{equation}
	where $\beta=\sqrt{6}\sigma/\pi$, $\alpha=\mu-\gamma\beta$, and $\gamma\approx0.5772$ is the Euler--Mascharoni constant. In addition, the loads are made mutually dependent through the C-vine copula with density
	\begin{equation}
	c_{\vX}^{(\mathcal G)}(u_{1},\ldots,u_{6})=\prod_{j=2}^{6}c_{1j; \theta=1.1}^{(\mathcal{GH})}(u_{1},u_{j}),
	\label{eq:TrussVine}
	\end{equation}
	where each $c_{1j; \theta}^{(\mathcal{GH})}$ is the density of the pair-copula between $P_{1}$ and $P_j$, $j=2,\ldots,\xdim$, and belongs to the Gumbel--Hougaard family defined in \tabref{pair_copula_cdfs}, row 11.
	
	The presence of the loads causes a downward vertical displacement $\Delta$ at the mid span of the structure. $\Delta$ is taken to be the system's uncertain response to the $10$-dimensional random input ${\vX=(E_1,E_2,A_1,A_2,P_1,\ldots,P_6)}$ consisting of the $4$ structural variables and the $6$ loads. The true statistics (mean, standard deviation, $\pdf$) of $\Delta$ are obtained by MCS over $10^7$ sample points, and are shown in the left panel of \figref{Ytruepdf_truss10d}.
	
	{
		\linespread{1}
		\begin{figure}[!ht]
			\begin{centering}
				\includegraphics[width=16cm]{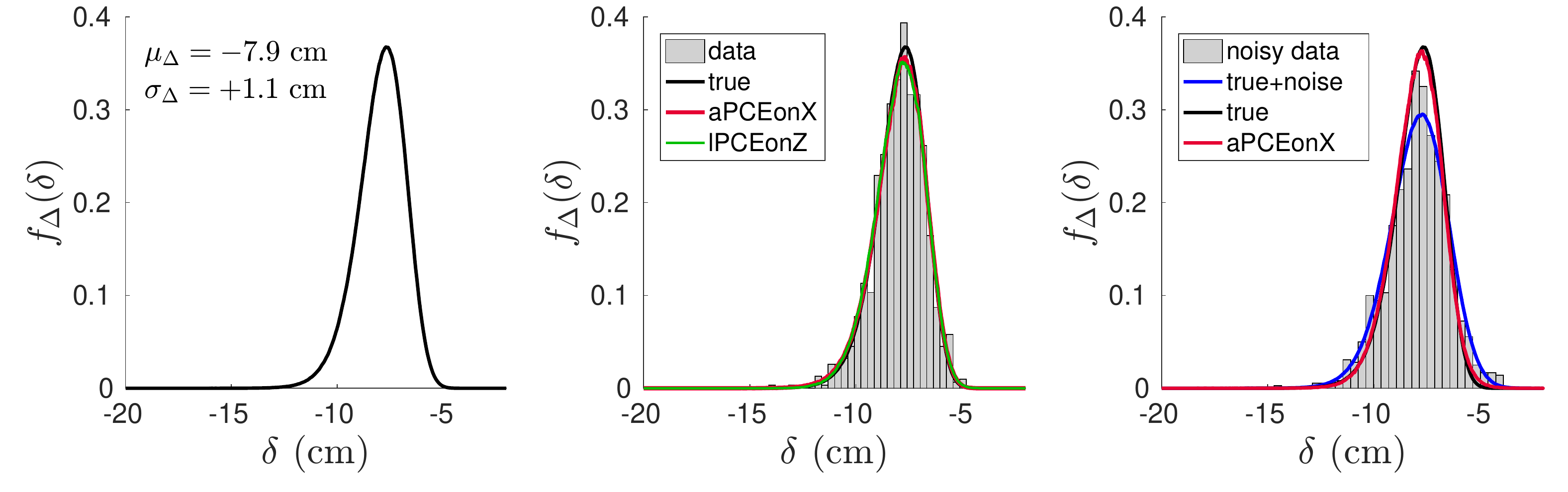}
				\caption{\textbf{Response PDFs of the horizontal truss.} \emph{Left panel:} true $\pdf$ $f_{Y}$ of the truss response, obtained on $10^7$ sample points by KDE. \emph{Central panel:} probability histogram obtained from $n'=1{,}000$ output observations used for training (gray bars), true response $\pdf$ as in the left panel (black), $\pdf$s obtained from the {\aPCEonX} (red) and the {\lPCEonZ} (green) by resampling. \emph{Right panel:} as in the central panel, but for training data perturbed with Gaussian noise ($\sigma_\varepsilon=0.79\cm=0.70\sigma_\Delta$). The blue line indicates the true $\pdf$ of the perturbed model.}
				\label{fig:Ytruepdf_truss10d}
			\end{centering}
		\end{figure}
		
		\begin{figure}[!ht]
			\begin{center}
				\includegraphics[width=16cm]{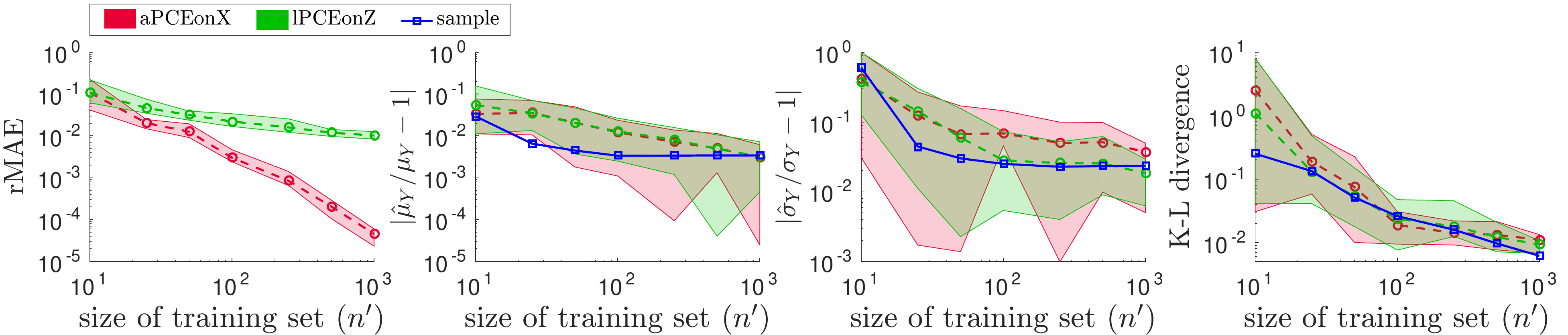}
			\end{center}
			\caption{\textbf{PCE performance for the horizontal truss}. From left to right: rMAE, error on the mean, error on the standard deviation, and Kullback-Leibler divergence of {\aPCEonX} (red) and {\lPCEonZ} (green), for a size $n'$ of the training set increasing from $10$ to $1{,}000$. The dash-dotted lines and the bands indicate, respectively, the average and the minimum to maximum errors over $10$ simulations. In the second to fourth panels, the blue lines correspond to the empirical estimates obtained from the training data (error bands not shown).}
			\label{fig:PCEperformance_Truss}
		\end{figure}
	}
	
	We analyse the system with the same procedure undertaken for the Ishigami model: we build {\aPCEonX} and {\lPCEonZ} on each of $10$ training sets $\dXYprime$ of increasing size $n'$, and validate their performance. The pointwise error is evaluated on $10$ validation sets $\dXYsecond$ of fixed size $n''=10{,}000$, while the statistical errors are determined by large resampling.
	
	The results are shown in \figref{PCEperformance_Truss}. Both PCEs exhibit high performance, yet the {\aPCEonX} yields a significantly smaller pointwise error (first panel). The {\lPCEonZ} yields a better estimate of the standard deviation, yet the empirical estimates obtained from the training data are the most accurate ones in this case.
	
	Having selected the {\aPCEonX} as the better of the two metamodels, we further assess its performance in the presence of noise. We perturb the response values used to train the model by adding Gaussian noise with increasing standard deviation $\sigma_\varepsilon$, set to $1\%$, $10\%$, and $30\%$ of $|\mu_\Delta|$ (equivalent to $7\%$, $70\%$, and $210\%$ of $\sigma_\Delta$, respectively). The results are shown in Figures \ref{fig:PCEperformance_noise_truss10d}-\ref{fig:PCEperformance_noise_truss10d_PCEvsMC}. The errors of all types are significantly smaller than the magnitude of the added noise, and decrease with increasing sample size for all noise levels (\figref{PCEperformance_noise_truss10d}). Also, the PCE estimates are significantly better than the sample estimates (\figref{PCEperformance_noise_truss10d_PCEvsMC}; see also \figref{Ytruepdf_truss10d}, fourth panel).
	
	{
		\linespread{1}
		\begin{figure}[!ht]
			\begin{center}
				\includegraphics[width=16cm]{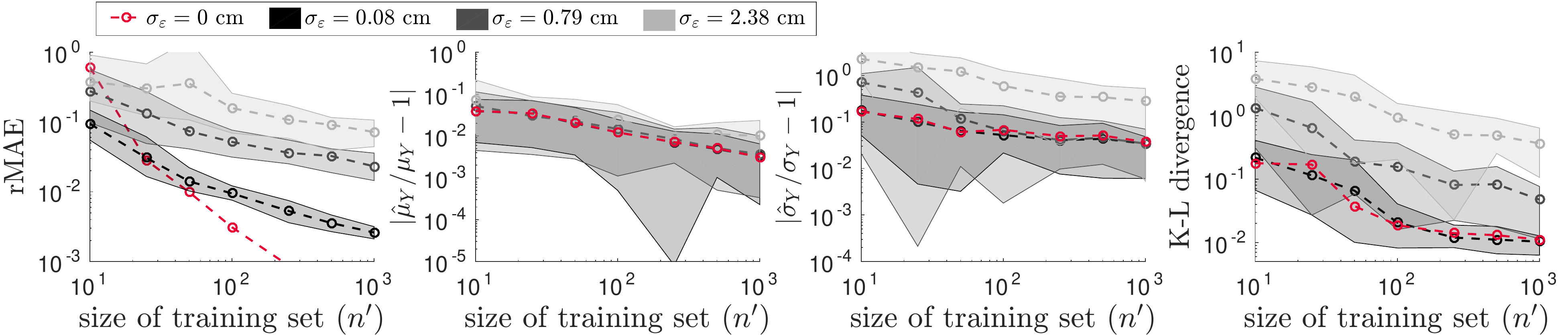}
			\end{center}
			\caption{\textbf{{\aPCEonX} of horizontal truss: robustness to noise (for multiple noise levels)}. From left to right: rMAE, error on the mean, error on the standard deviation, and Kullback-Leibler divergence of {\aPCEonX} for an increasing amount of noise: $\sigma_\varepsilon=0.079\cm$ (dark gray), $\sigma_\varepsilon=0.79\cm$ (mid-light gray), and $\sigma_\varepsilon=2.38\cm$ (light gray). The dash-dotted lines and the bands indicate, respectively, the average and the minimum to maximum error over $10$ simulations. The red lines, reported from \figref{PCEperformance_Truss} for reference, indicate the error for the noise-free data.}
			\label{fig:PCEperformance_noise_truss10d}
		\end{figure}
		
		\begin{figure}[!ht]
			\begin{center}
				\includegraphics[width=16cm]{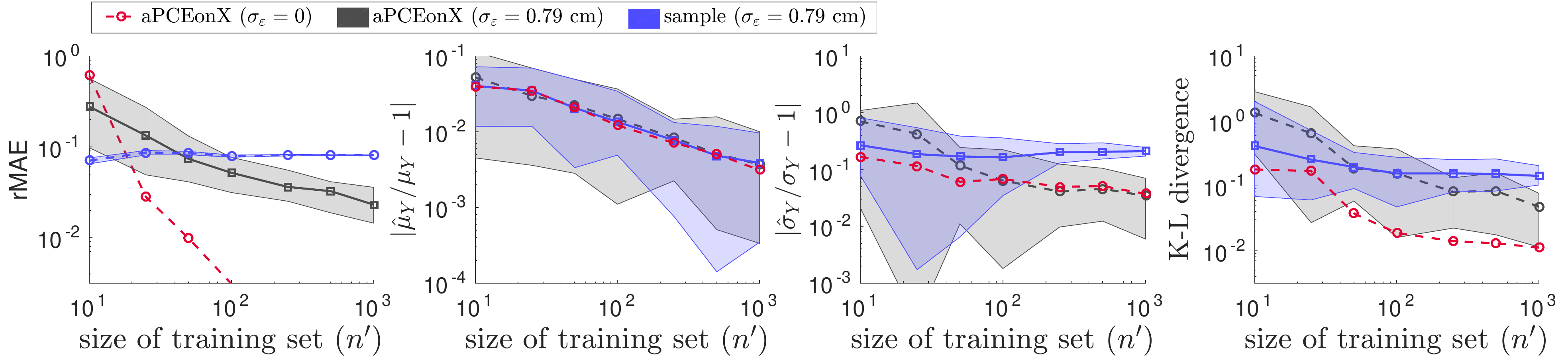}
			\end{center}
			\caption{\textbf{{\aPCEonX} of horizontal truss: robustness to noise (w.r.t. sample estimation)}. From left to right: rMAE, error on the mean, error on the standard deviation, and Kullback-Leibler divergence obtained by {\aPCEonX} (gray) and by direct sample estimation (blue), for noise ${\sigma_\varepsilon=0.79\cm=0.7\,\sigma_Y}$. The dash-dotted lines and the bands indicate, respectively, the average and the minimum to maximum error over $10$ simulations. The red lines, reported from \figref{PCEperformance_synthetic_models} for reference, indicate the mean error obtained for the noise-free data.}
			\label{fig:PCEperformance_noise_truss10d_PCEvsMC}
		\end{figure}
	}

	\subsection{Regularization without orthonormalization} \label{subsec:PCEvsSALSA}
	
	As discussed in the \subsecref{PCE}, PCE first determines a basis of polynomials which are mutually orthonormal with respect to the input PDF, then expresses the system response as a sum of such polynomials. The expansion coefficients are determined, in large dimension $M$ and in the examples above, by sparse regression. Here in particular, LAR was selected as a method to solve sparse regression.
	
	In the presence of correlations among the inputs, we further explored two strategies. The first one, \lPCEonZ, transforms the inputs into statistically independent random variables, and then builds an orthonormal basis with respect to the new variables. The second strategy, \aPCEonX, ignores the input dependencies (copula) and builds a basis orthonormal with respect to the product of their marginals. \aPCEonX{} ultimately outperforms \lPCEonZ{} in terms of pointwise accuracy, because it avoids highly non-linear transformations that make the expansion spectrum decay slower and thus the polynomial representation less efficient.
	
	An even simpler alternative to \aPCEonX{} would be to ignore not only the input copula, but its marginal distributions as well, and to use any non-orthonormal polynomial basis. Various parametric and non-parametric schemes exist to this end. We consider three different ones, which we apply to the Ishigami and horizontal truss models. This investigation allows us to quantify the benefit of basis orthonormality.
	
	The first of such methods is LASSO \cite{Tibshirani1996_LASSO}, which directly solves the problem \eqref{eq:LAR}. In standard LASSO the complete basis comprises all and only monomials of the type $\prod_{i=1}^{M} x_i^{k_i}$, $k_i \in \mathbb{N}^+$. In practice though, the number of different monomials is reduced by neglecting interactions, that is, by considering only monomials of the type $x_i^{k_i}$. Further simplification is typically achieved by considering only the linear terms. It is clear that, in such settings, LASSO would miss the complex interactions that characterize both the Ishigami and the horizontal truss systems. Indeed, using the standard Matlab implementation of linear LASSO (with $10$-fold cross-validation), we could not achieve a better accuracy  than $\rMAE=0.05$ for either case.
	
\begin{figure}
		\begin{center}
			\includegraphics[width=5cm]{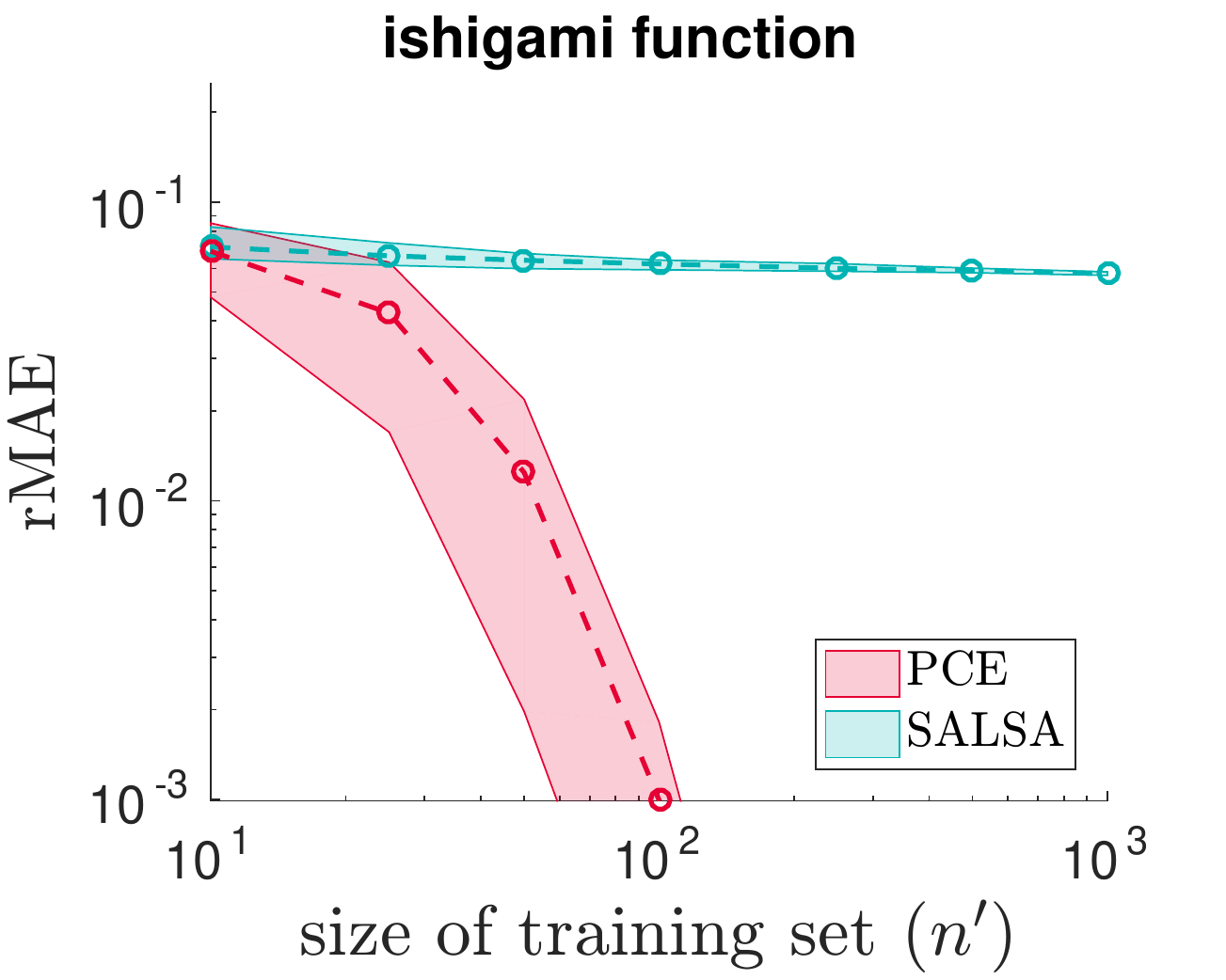}
			\includegraphics[width=5cm]{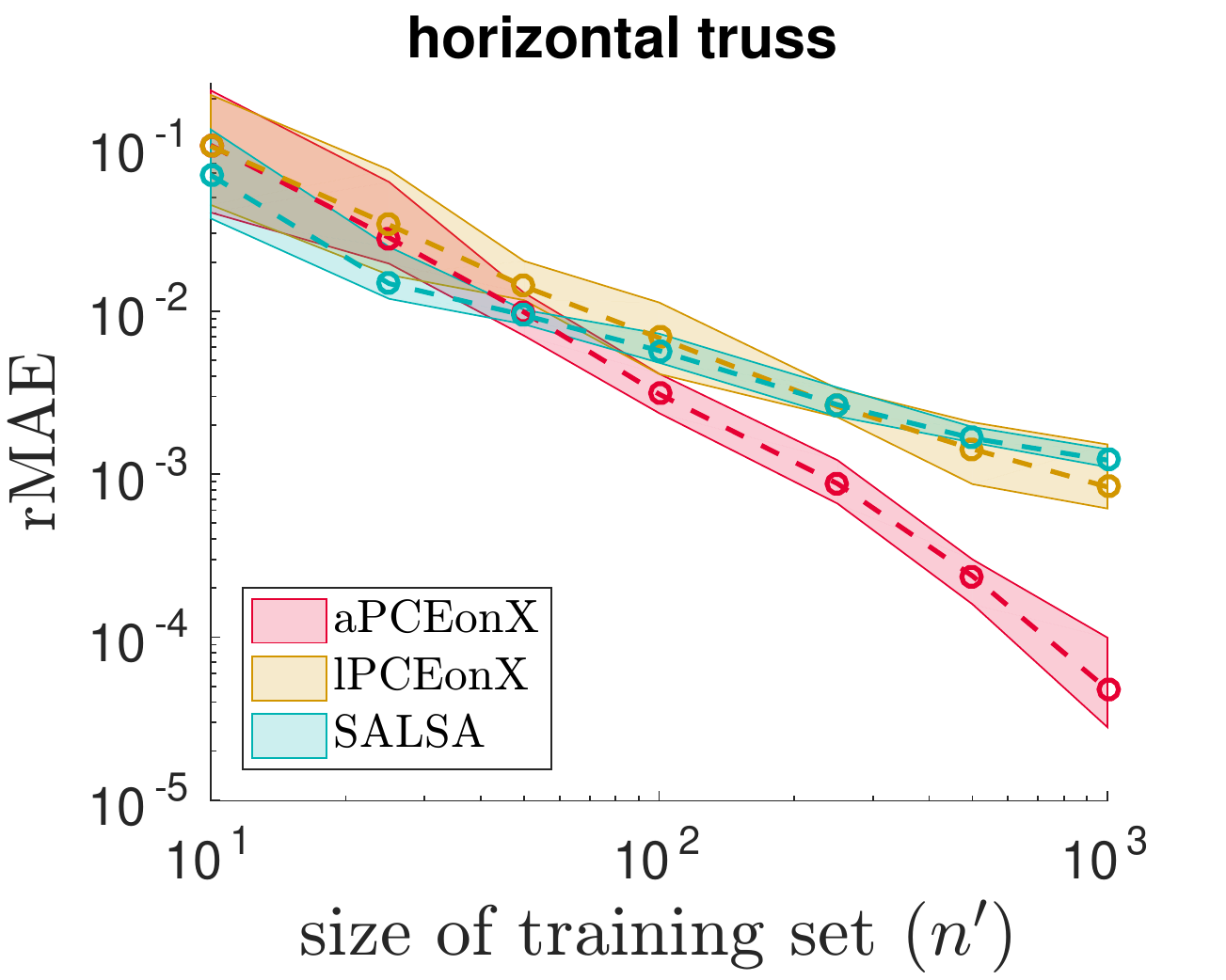}		
		\end{center}
		\caption{\textbf{PCE vs SALSA analysis of Ishigami and horizontal truss data.} rMAE of \aPCEonX{} (red) vs SALSA (cyan) on the Ishigami (left) and horizontal truss (right) data, for increasing training set size.}
		\label{fig:results_SALSA}
	\end{figure}

	A recent improvement of LASSO is SALSA \cite{Kandasamy2016_SALSA}. The method considers interactions among the input variables, and automatically selects the relevant terms in the expansion by an approach analogous to kernel ridge regression. We analyse the Ishigami and horizontal truss data with SALSA, using the Matlab implementation provided by the authors at \url{github.com/kirthevasank/salsa}. The results are illustrated in \figref{results_SALSA}. While performing overall better than LASSO, SALSA (cyan curve) fails to capture the highly non-linear behavior of the Ishigami function, and performs overall worse than \aPCEonX{} (red) both on the Ishigami and the horizontal truss models.
	
	The third attempt to solve sparse regression using a non-orthonormal basis is by performing PCE on the data $\dX$ using Legendre polynomials. We refer to this approach as \lPCEonX{}. Legendre polynomials are the family of orthonormal polynomials with respect to uniform distributions in an interval $[a,b]$. We apply \lPCEonX{} to the horizontal truss model only, as for the Ishigami model - whose inputs are truly uniform - it would correspond to the non-interesting case of assuming the correct marginals. We obtain the interval $[a,b]$ for each input variable from the training data, as the observation range enlarged by $1\%$ both to the left and to the right. The results are shown in \figref{results_SALSA}, right, orange curve. \lPCEonX{} performs similarly to SALSA, and consistently worse than \aPCEonX.

	These comparisons demonstrate numerically the benefit of performing regression using an orthonormal basis, which ensures, due to finite variance of the response, better compression and thus more rapid convergence.
	
	\subsection{Preliminary conclusion} \label{subsec:validation_conclusions}
	
	The results obtained in the previous section allow us to draw some important preliminary conclusions on data-driven PCE. The methodology:
	\begin{itemize} \setlength{\itemsep}{0em}
		\item delivers reliable predictions of the system response to multivariate inputs;
		\item produces reliable estimates of the response statistics if the input dependencies are properly modelled, as done here through copulas (for {\aPCEonX}: a-posteriori);
		\item works well already when trained on few observations;
		\item deals effectively with noise, thus providing a tool for denoising;
		\item involves only few hyperparameters, which have a clear interpretation and are easily tuned by cross-validation;
		\item outperforms sparse regression schemes based on non-orthonormal bases, such as LASSO \cite{Tibshirani1996_LASSO}, SALSA \cite{Kandasamy2016_SALSA}, or \lPCEonX{} (see previous section).
	\end{itemize}
	
	In order to build the expansion when the inputs are mutually dependent, we investigated two alternative approaches, labelled as {\lPCEonZ} and {\aPCEonX}. Of the two strategies, {\aPCEonX} appears to be the most effective one in purely data-driven problems. It is worth mentioning, though, that {\lPCEonZ} may provide superior statistical estimates if the joint distribution of the input is known with greater accuracy than in the examples shown here. This was the case, for instance, when we replaced the inferred marginals and copula used to build the {\lPCEonZ} with the true ones (not shown here): in both examples above, we obtained more accurate estimates of $\mu_Y$, $\sigma_Y$, and $F_Y$ (but not better pointwise predictions) than using {\aPCEonX}. 
	
	The simulation results show a mismatch between the fast speed of MAE convergence and the slower convergence speed of the other metrics. Such mismatch is explained by the data-driven setup. To see this, let us start from an extreme example. Consider the ideal case where the true model is an exact polynomial. Inferring the input joint $\pdf$ $\fX$ from data (first step of the analysis) will generally lead to an inferred pdf $\hat{\fX} \not\equiv \fX$. The bases associated to $\fX$ and $\hat{\fX}$ will thus also differ. Nevertheless, performing regression on the basis associated to $\hat{\fX}$ may still lead to an exact expansion, that is, to a PCE model that is identical to the true polynomial. In this ideal scenario, the MAE would be $0$. Yet, estimating the output statistics would involve resampling input points from the wrong pdf $\hat{\fX}$. The error would be propagated by the correct PCE, and yield positive errors in the statistical estimates despite the PCE model being exact. The numerical examples shown here are not so extreme: the original model $\MODEL$ is not a polynomial and the obtained PCE only approximates $\MODEL$. Still, it is a good approximation, in the sense that the resulting MAE is very small even if the wrong input $\pdf$ (and therefore a basis which is not exactly orthonormal) is used. The output statistics instead have a larger estimation error, because their accuracy is more strongly affected by errors in the inferred input distribution.

	\section{Results on real data sets} \label{sec:applications}
	
	We now demonstrate the use of {\aPCEonX} on three different real data sets. The selected data sets were previously analysed by other authors with different machine learning algorithms, which establish here the performance benchmark. 
	
	\subsection{Analysis workflow}
	
	\subsubsection{Statistical input model} \label{subsec:statistical_input_models}
	
	The considered data sets comprise samples made of multiple input quantities and one scalar output. Adopting the methodology outlined in \secref{methods}, we characterize the multivariate input $\vX$ statistically by modelling its marginals $\hat{f}_i$ through KDE, and we then resort to arbitrary PCE to express the output $Y$ as a polynomial of $\vX$. The basis of the expansion thus consists of mutually orthonormal polynomials with respect to $\prod_i \hat{f}_i(x_i)$, where $\hat{f}_i$ is the marginal $\pdf$ inferred for $X_i$. 
	
	\subsubsection{Estimation of pointwise accuracy}  \label{subsec:validation}
	
	Following the pointwise error assessment procedure carried out in the original publications, for the case studies considered here we assess the method's performance by cross-validation. Standard $k$-fold cross-validation partitions the data $\dXY$ into $k$ subsets, trains the model on $k-1$ of those (the training set $\dXYprime$), and assesses the pointwise error between the model's predictions and the observations on the $k$-th one (the validation set $\dXYsecond$). The procedure is then iterated over all $k$ possible combinations of training and validation sets. The final error is computed as the average error over all validation sets. The number $k$ of data subsets is chosen as in the reference studies. Differently from the synthetic models considered in the previous section, the true statistics of the system response are not known here, and the error on their estimates cannot be assessed.
	
	A variation on standard cross-validation consists in performing a $k$-fold cross validation on each of multiple random shuffles of the data. The error is then typically reported as the average error obtained across all randomisations, ensuring that the final results are robust to the specific partitioning of the data in its $k$ subsets. In the following, we refer to a $k$-fold cross validation performed on $r$ random permutations of the data (i.e. $r$ random $k$-fold partitions) as an $r \times k$-fold randomised cross-validation.
	
	\subsubsection{Statistical estimation} \label{subsubsec:realdata_APestimation}
	
	Finally, we estimate for all cases studies the response $\pdf$ a-posteriori (AP) by resampling. To this end, we first model their dependencies through a C-vine copula $\hat{C}^{(\mathcal V)}$. The vine is inferred from the data as detailed in \subappref{Vine_construction_in_practice}. Afterwards, resampling involves the following steps:
	\begin{itemize} \setlength{\itemsep}{0em}
		\item sample $n_{\AP}$ points $\dZ_{\AP}=\{\vz^{(l)}, l=1,\ldots,n_{\AP}\}$ from $\vZ \sim U([0,1])^d$. We opt for Sobol' quasi-random low-discrepancy sequences \cite{Sobol1967_86}, and set $n_{\AP}=10^6$;
		\item map $\dZ_{\AP} \mapsto \dU_{\AP} \subset [0,1]^d$ by the inverse Rosenblatt transform of $\hat{C}^{(\mathcal V)}$;
		\item map $\dU_{\AP} \mapsto \dX_{\AP}$ by the inverse probability integral transform of each marginal $\cdf$ $\hat{F}_i$. $\dX_{\AP}$ is a sample set of input observations with copula $\hat{C}^{(\mathcal V)}$ and marginals $\hat{F}_i$;
		\item evaluate the set $\dY_{\AP}=\{ y_{\PC}^{(l)}=\MODEL_{\PC}(\vx), \, \vx \in \dX_{\AP} \}$ of responses to the inputs in $\dX_{\AP}$.
	\end{itemize}
	The $\pdf$ of $Y$ is estimated on $\dY_{\AP}$ by kernel density estimation.

	\subsection{Combined-cycle power plant} \label{subsec:CCPP}
	
	The first real data set we consider consists of $9{,}568$ data points collected from a combined-cycle power plant (CCPP) over $6$ years (2006-2011). The CCPP generates electricity by gas and steam turbines, combined in one cycle. The data comprise $4$ ambient variables and the energy production $E$, measured over time. The four ambient variables are the temperature $T$, the pressure $P$ and the relative humidity $H$ measured in the gas turbine, and the exhaust vacuum $V$ measured in the steam turbine. All five quantities are hourly averages. The data are available online \cite{Lichman2013}. 
	
	The data were analysed in \cite{Tuefekci2014_126} with $13$ different ML techniques, including regression trees, different types of neural networks (NNs), and support vector regression (SVR), to predict the energy output based on the measured ambient variables. The authors assessed the performance of each method by $5\times 2$-fold randomised cross-validation, yielding a total of $10$ pairs of training and validation sets. Each set contained $4{,}784$ instances. The best learner among those tested by the authors was a bagging reduced error pruning (BREP) regression tree. Specifically, the model was obtained as the average response (``bagging'') of $10$ pruned regression trees, each one built on a subset of the full data obtained by bootstrapping. Each pruned regression tree was obtained as a regression tree that was then iteratively simplified by reduced error pruning. The final model yielded a mean MAE of $3.22\MWh$ (see their Table 10, row 4). The lowest MAE of this model over the $10$ validation sets, corresponding to the ``best'' validation set, was indicated to be $2.82\MWh$. Besides providing an indicative lower bound to the errors, the minimum gives, when compared to the means, an indication of the variability of the performance over different partitions of the data. The actual error variance over the $10$ validation sets was not provided in the mentioned study.
	
	\begin{table}
		\linespread{1}
		\begin{center}
			\begin{tabular}{rcccc}
				\toprule
				& MAE & min. MAE & mean-min & rMAE ($\%$) \\
				\cmidrule{2-5} 
				{\aPCEonX}    				    & 3.11 $\pm$ 0.03 & 3.05 & 0.06 & 0.68 $\pm$ 0.007 \\
				BREP \cite{Tuefekci2014_126} & 3.22 $\pm$ n.a. & 2.82 & 0.40 & n.a. \\
				\bottomrule
			\end{tabular}
			\caption{\textbf{Errors on CCPP data.} MAE yielded by the {\aPCEonX} (first row) and by the BREP regression tree in \cite{Tuefekci2014_126} (second row). From left to right: average MAE $\pm$ its standard deviation over all $10$ validation sets (in $\MWh$), its minimum (error on the ``best set''), difference between the average and the minimum MAEs, and rMAE.}
			\label{tab:CCPP_results}
		\end{center}
	\end{table} 
	
	We analyze the very same $10$ training sets by PCE. The results are reported in \tabref{CCPP_results}. The average MAE yielded by the {\aPCEonX} is slightly smaller than that of the BREP model. More importantly, the difference between the average and the minimum error, calculated over the $10$ validation sets, is significantly lower with our approach, indicating a lower sensitivity of the results to the partition of the data, and therefore a higher reliability in the presence of random observations. The average error of the PCE predictions relative to the observed values is below $1\%$. 
	
\begin{figure}[!ht]
		\linespread{1}
		\begin{center}
			\includegraphics[width=10cm]{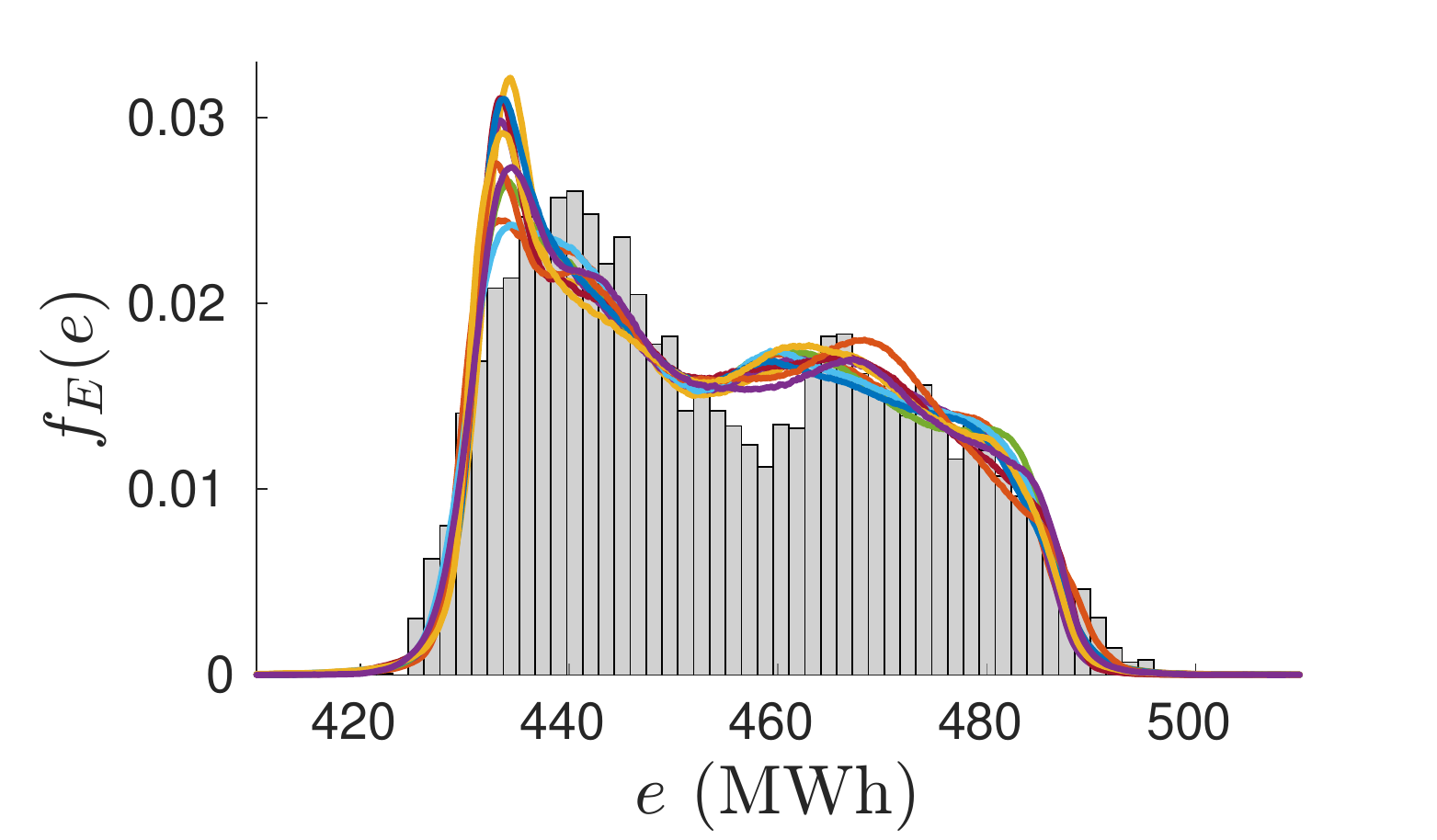}
		\end{center}
		\caption{\textbf{Estimated $\pdf$ of the energy produced by the CCPP.} The bars indicate the histogram obtained from the observed CCPP energy output. The coloured lines show the $\pdf$s of the PCE metamodels built on the $10$ training sets, for input dependencies modelled by C-vines.}
		\label{fig:CCPP_Ypdf}
	\end{figure}

	Finally, we estimate the $\pdf$ of the hourly energy produced by the CCPP following the procedure described in \subsubsecref{realdata_APestimation}. The results are shown in \figref{CCPP_Ypdf}. Reliable estimates of the energy $\pdf$ aid for instance energy production planning and management.
	
	\subsection{Boston Housing} \label{subsec:boston_housing}
	
	The second real data set used to validate the PCE-based ML method concerns housing values in the suburbs of Boston, collected in 1970. The data set, downloaded from \cite{Lichman2013}, was first published in \cite{Harrison1978_BostonHousing}, and is a known reference in the machine learning and data mining communities. 
	
	The data comprise $506$ instances, each having $14$ attributes. One attribute (the proximity of the neighborhood to the Charles river) is binary-valued and is therefore disregarded in our analysis. Of the remaining $13$ attributes, one is the median housing value of owner-occupied homes in the neighbourhood, in thousands of $\$$ (MEDV). The remaining $12$ attributes are, in order: the per capita crime rate by town (CRIM), the proportion of residential land zones for lots over 25,000 sq.ft. (ZN), the proportion of non-retail business acres per town (INDUS), the nitric oxides concentration, in parts per 10 million (NOX), the average number of rooms per dwelling (RM), the proportion of owner-occupied units built prior to 1940 (AGE), the weighted distances to five Boston employment centres (DIS), the index of accessibility to radial highways (RAD), the full-value property-tax rate per $\$$10,000 (TAX), the pupil-teacher ratio by town (PTRATIO),  the index $1{,}000(\textrm{Bk} - 0.63)^2$, where $\textrm{Bk}$ is the proportion of black residents by town, and the lower status of the population (LSTAT). 
	
	The data were analysed in previous studies with different regression methods to predict the median house values MEDV on the basis of the other attributes \cite{Can1992_BostonHousing, Gilley1995_BostonHousing, Quinlan1993_BostonHousing, KelleyPace1997_BostonHousing}. The original publication itself \cite{Harrison1978_BostonHousing} was concerned with determining whether the demand for clean air affected housing prices. The data were analysed with different supervised learning methods in \cite{Quinlan1993_BostonHousing}. Among them, the best predictor was shown to be an NN model combined with instance-based learning. The NN consisted of a single hidden layer and an output unit. The weights were regularised by a penalty coefficient. The number of hidden units and the penalty coefficients were first optimised by three-fold cross validation on the first training set, and each weight was then finally selected as the best of $3$ values randomly sampled within $[-0.3, +0.3]$. The final NN yielded $\MAE=2{,}230\$$ (rMAE: $12.9\%$) on a $10$-fold cross-validation. 
	
\begin{figure}[!ht]
		\linespread{1}
		\begin{center}
			\includegraphics[width=10cm]{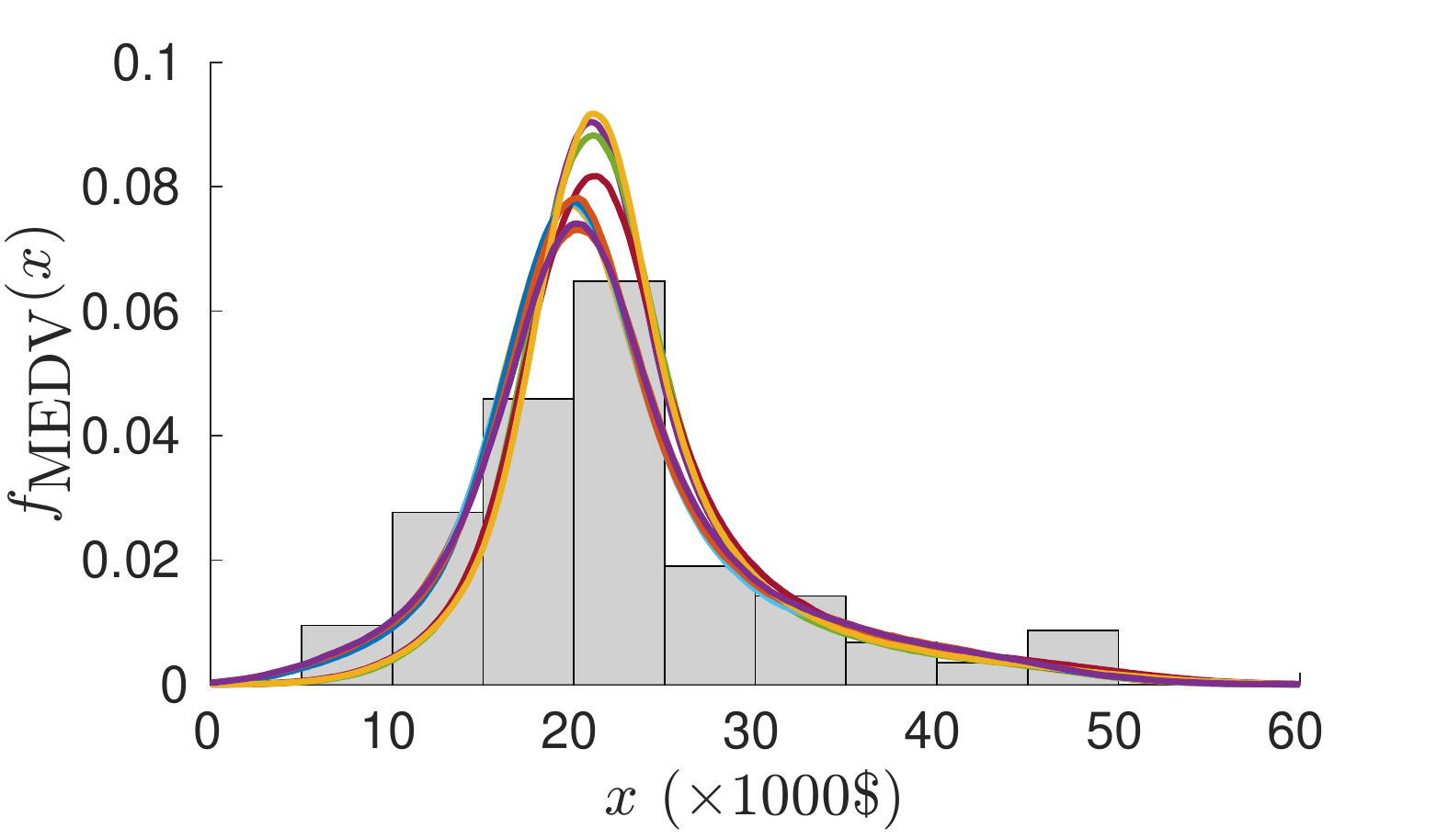}
		\end{center}
		\caption{\textbf{Estimated $\pdf$ of the variable MEDV.} The bars indicate the sample $\pdf$, as an histogram obtained using $50$ bins from $\$0$ to $\$50$k (the maximum house value in the data set). The coloured lines show the $\pdf$s of the PCE metamodels built on $10$ of the $100$ training sets (one per randomisation of the data), for input dependencies modelled by C-vines. The dots indicate the integrals of the estimated $\pdf$s for house values above $\$49$k. }
		\label{fig:BH_Ypdf}
	\end{figure}

	We model the data by PCE and quantify the performance by $10\times10$ randomised cross-validation. The results are summarised in \tabref{BH_results}. The errors are comparable to the NN model with instance-based learning in \cite{Quinlan1993_BostonHousing}. While the latter yields the lowest absolute error, the {\aPCEonX} achieves a smaller relative error. In addition, it does not require the fine parameter tuning that affects most NN models. Finally, we estimate the $\pdf$ of the median house value as described in \subsubsecref{realdata_APestimation}. The results are shown in \figref{BH_Ypdf}.
	
	\begin{table}[!ht]
		\linespread{1}
		\begin{center}
			\begin{tabular}{rcc}
				\toprule
				& MAE (\$)        & rMAE ($\%$) \\
				\cmidrule{2-3} 
				{\aPCEonX}    				        & 2483 $\pm$ 337  & 12.6 $\pm$ 2.0 \\
				NN \cite{Quinlan1993_BostonHousing} & 2230 $\pm$ n.a. & 12.9 $\pm$ n.a. \\
				\bottomrule
			\end{tabular}
			\caption{\textbf{Errors on Boston Housing data.} MAE and rMAE yielded by the {\aPCEonX} (first row) and by the NN model with instance-based learning from \cite{Quinlan1993_BostonHousing} (second row).}
			\label{tab:BH_results}
		\end{center}
	\end{table} 
	
	\subsection{Wine quality} \label{subsec:wine_quality}
	
	The third real data set we consider concerns the quality of wines from the \emph{vinho verde} region in Portugal. The data set consists of $1{,}599$ red samples and $4{,}898$ white samples, collected between 2004 and 2007. The data are available online at \url{http://www3.dsi.uminho.pt/pcortez/wine/}. Each wine sample was analysed in laboratory for $11$ physico-chemical parameters: 
	fixed acidity
	, volatile acidity
	, citric acid
	, residual sugar
	, chlorides
	, free sulfur dioxide
	, total sulfur dioxide
	, density
	, pH%
	, sulphates
	, alcohol
	. In addition, each sample was given a quality score $Q$ 
	based on blinded sensory tests from three or more sensory assessors. The score is the median of the grades (integers between 0 and 10) assigned by each assessor.
	
	The data were previously analysed in \cite{Cortez2009_547} to predict the wine quality score on the basis of the $11$ physico-chemical parameters. The study compared various ML algorithms, namely multiple regression, a single-layer NN, and linear support vector machine (SVM, \cite{Scholkopf2002}) regression. The NN consisted of a single hidden layer with logistic transfer function and an output node with linear function. The number of hidden nodes and the weights were automatically calibrated by cross-validation. For linear SVM regression, a Gaussian kernel was selected, with kernel width tuned by cross-validation. The remaining parameters were fixed to arbitrary values to reduce the computational cost. SVM regression outperformed the other methods (at the expense of a considerably higher computational cost), yielding the lowest MAE, as assessed by means of $20\times$ $5$-fold randomised cross-validation.
	
	We model the data by {\aPCEonX}, and round the predicted wine ratings, which take continuous predicted values, to their closest integer. The performance is then assessed through the same cross-validation procedure used in \cite{Cortez2009_547}. The results are reported in \tabref{wine_results}. The MAE of {\aPCEonX} is comparable to that of the SVM regressor, and always lower than the best NN. 
	
	\begin{table}[!ht]
		\linespread{1}
		\begin{center}
			\begin{tabular}{rcccccc}
				\toprule
				{ } & {\quad} & \multicolumn{2}{l}{red wine:} & { \qquad } & \multicolumn{2}{l}{white wine:} \\[-6pt]
				{ } & { }      & MAE  & rMAE ($\%$)  & {} & MAE  & rMAE ($\%$)  \\
				\cmidrule{3-4} \cmidrule{6-7}
				{\aPCEonX}               & { }	& 0.44 $\pm$ 0.03  &  8.0 $\pm$ 0.6 &
				& 0.50 $\pm$ 0.02 & 8.8 $\pm$ 0.3 \\
				SVM in \cite{Cortez2009_547}    &  & 0.46 $\pm$ 0.00  & n.a.  & 
				& 0.45 $\pm$ 0.00 & n.a. \\
				Best NN in \cite{Cortez2009_547} & & 0.51 $\pm$ 0.00 & n.a. & 
				& 0.58 $\pm$ 0.00 & n.a. \\[8pt]
				\bottomrule
			\end{tabular}
			\caption{\textbf{Errors on wine data.} MAE and rMAE yielded on red and white wine data by the {\aPCEonX}, by the SVM in \cite{Cortez2009_547}, and by the best NN model in \cite{Cortez2009_547}.}
			\label{tab:wine_results}
		\end{center}
	\end{table} 
	
\begin{figure}[!ht]
		\linespread{1}
		\begin{center}
			\includegraphics[width=16cm]{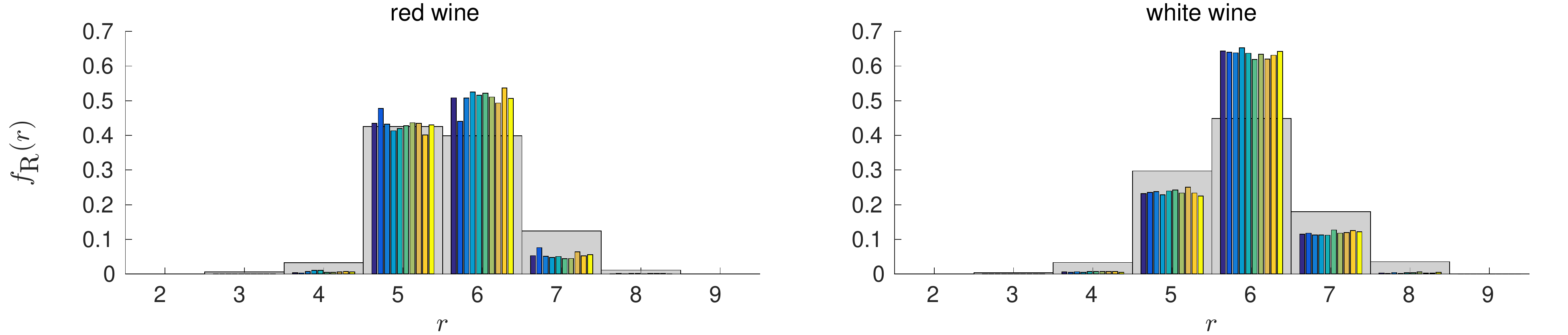}
		\end{center}
		\caption{\textbf{Estimated $\pdf$ of the wine rating.} For each panel (left: red wine; right: white wine), the grey bars indicate the sample $\pdf$ of the median wine quality score assigned by the assessors. The coloured bars show the predicted $\pdf$s obtained by resampling from the PCE metamodels built on $10$ of the $100$ total training sets, for input dependencies modelled by C-vines.}
		\label{fig:Wine_Ypdf}
	\end{figure}

	Finally, our framework enables the estimation of the $\pdf$ of the wine rating as predicted by the PCE metamodels. The resulting $\pdf$s are shown in \figref{Wine_Ypdf}. One could analogously compute the conditional $\pdf$s given by fixing any subset of inputs  to given values (\eg, the residual sugar or alcohol content, which can be easily controlled in the wine making process). This may help, for instance, predicting the wine quality for fixed physico-chemical parameters, or choosing the latter so as to optimize the wine quality or to minimize its uncertainty. This analysis goes beyond the scope of the present work. 
	
	\section{Discussion and conclusions} \label{sec:discussion}
	
	We proposed an approach to machine learning (ML) that capitalises on polynomial chaos expansion (PCE), an advanced regression technique from uncertainty quantification. PCE is a popular spectral method in engineering applications, where it is often used to replace expensive-to-run computational models subject to uncertain inputs with an inexpensive metamodel that retains the statistics of the output (\eg, moments, $\pdf$). Our paper shows that PCE can also be used as an effective regression model in purely data-driven problems, where only input observations and corresponding system responses - but no computational model of the system - are available. This setup poses various challenges that PCE does not have to deal with in classical uncertainty quantification problems. Namely, the training set is fixed and cannot be enriched, the input joint distribution (needed to build the spectral basis and to draw statistical estimates) is unknown and has to be entirely inferred from the available data, and the data may be noisy.
	
	We tested the performance of PCE on simulated data first, and then on real data by cross-validation. The reference performance measure was the mean absolute error of the PCE metamodel over all test data. The simulations also allowed us to assess the ability of PCE to estimate the statistics of the response (its mean, standard deviation, and $\pdf$) in the considered data-driven scenario. Both the pointwise and the statistical errors of the methodology were low, even when relatively few observations were used to train the model. Importantly, high performance was still obtained in the presence of strong noise in the simulated data. PCE can thus be used for denoising, a feature that had not been previously investigated. Finally, the numerical experiments demonstrated the superiority of PCE, which uses as expansion basis the set of orthonormal polynomials with respect to the product of the input marginals, over methods that perform sparse regression over a non-orthonormal basis, such as LASSO \cite{Tibshirani1996_LASSO}, SALSA \cite{Kandasamy2016_SALSA}, or PCE built enforcing a different basis.
	
	The applications to real data showed a comparable, and sometimes slightly superior, performance to that of other ML methods used in previous studies, such as different types of neural networks and support vector machines. In general, the quality of the approximations will depend on the magnitude of the truncation error, which is dictated by the speed of the decay of the spectrum. The amount of data available also limits in practice the attainable expansion order, as already observed in \cite{Oladyshkin2018_137}.
	
	PCE offers several advantages over other established ML regression techniques. First, the framework performs well on very different tasks, with only little parameter tuning needed to adapt the methodology to the specific data considered. The quadratic nature of the PCE regression problem enables direct optimisation, compared for instance to the cumbersome iterative training/calibration procedure in NNs. For PCE, only the total degree $p$, the $q$-norm parameter and the interaction degree $r$ are to be specified. Automatic calibration of these parameters within pre-assigned ranges is straightforward. Specifically, as a single analysis takes only a few seconds to a minute to be completed on a standard laptop (depending on the size of the data), it is possible to repeat the analysis over an array of allowed $(p,q,r)$ values, and to retain the PCE with minimum error in the end. In our analyses we tuned the parameters $p$ and $r$ in this way, while setting $q=0.75$ (not optimized, in order to reduce the total computational cost). This feature distinguishes PCE from the above mentioned ML methods, which instead are known to be highly sensitive to their hyperparameters and require an appropriate and typically time consuming calibration \cite{Claesen2015_arxiv}. Indeed, it is worth noting that, in the comparisons we made, all PCE metamodels were built by using the very same procedure, including the automatic hyperparameter calibration. When compared to the best NNs or SVM found in other studies, which differed significantly from each other in their construction and structure, the PCE metamodels exhibited a comparable performance. 
	
	Second, PCE delivers not only accurate pointwise predictions of the output for any given input, but also statistics thereof in the presence of input uncertainties. This is made possible by combining the PCE metamodel with a proper probabilistic characterisation of the input uncertainties through marginal distributions and copulas, and then by resampling from the obtained input model. The result is not trivial: theory motivates the spectral convergence of PCE in classical UQ settings (see \subsecref{PCE_representation}). In the data-driven setup considered here, however, the joint $\pdf$ of the input is unknown and has to be inferred (see also \subsecref{validation_conclusions}), a step which generally harms convergence. As a further difficulty, we addressed the worst-case scenario where parametric inference is not possible due to missing information, and proceeded instead by kernel density estimation. Despite these challenges, quite accurate statistics were obtained. The methodology works well also in the presence of several inputs (as tested on simulated problems of dimension up to $10$) and of sample sets of comparably small size. It is worth mentioning in this respect that the same resampling approach could be applied in combination with any technique for function approximation (including neural networks and support vector machines). In the particular case of independent inputs, the PCE representation encodes the moments of the output and its Sobol sensitivity indices directly in the polynomial coefficients \cite{SudretRESS2008b}.
	
	Third, the analytical expression of the output yielded by PCE in terms of a simple polynomial of the input makes the model easy to interpret, compared for instance to other popular methods for ML regression.
	
	Fourth, the polynomial form makes the calibrated metamodel portable to embedded devices (\eg, drones). For this kind of applications, the demonstrated robustness to noise in the data is a particularly beneficial feature. 
	
	Fifth and last, PCE needs relatively few data points to attain acceptable performance levels, as shown here on various test cases. This feature demonstrates the validity of PCE metamodelling for problems affected by data scarcity, also when combined with complex vine copula representations of the input dependencies.
	
	One limitation of PCE-based regression as presented here is its difficulty in dealing with data of large size or consisting of a very large number of inputs. Both features lead to a substantially increased computational cost needed to fit the PCE parameters and (if statistical estimation is wanted and the inputs are dependent) to infer the copula. Various solutions can be nevertheless envisioned. In the presence of very large training sets, the PCE may be initially trained on a subset of the available observations, and subsequently refined by enriching the training set with points in the region where the observed error is larger. Regarding copula inference, which is only needed for an accurate quantification of the prediction's uncertainty, a possible solution is to employ a Gaussian copula. The latter involves a considerably faster fitting than the more complex vine copulas, and still yielded in our simulations acceptable performance. Alternatively, one may reduce the computational time needed for parameter estimation by parallel computing, as done in \cite{Wei2016_arXiv}. Recently, a technique to efficiently infer vine copulas in high-dimensional problems has been proposed in \cite{Mueller2018_VineHighDim}.
	
	Finally, the proposed methodology has been shown here on data characterised by continuous input variables only. PCE construction in the presence of discrete data is equally possible, and the Stiltjes orthogonalisation procedure is known to be quite stable in that case \cite{Gautschi1982_289}. The a-posteriori quantification of the output uncertainty, however, generally poses a challenge. Indeed, it involves the inference of a copula among discrete random variables, which requires a different construction \cite{Genest2007_475}. Recently, however, methods have been proposed to this end, including inference for R-vines \cite{Panagiotelis2012_1063, Panagiotelis2017_138}. Further work is foreseen to integrate these advances with PCE metamodelling.

	\section*{Acknowledgments} 
	
	Emiliano Torre gratefully acknowledges financial support from RiskLab, Department of Mathematics, ETH Zurich and from the Risk Center of the ETH Zurich. Declarations of interest: none.
	
	\bibliographystyle{chicago} 
	\bibliography{DataDrivenPCE_bibliography}
	
	\appendix
	
	\setcounter{equation}{0}
	\renewcommand\thetable{\thesection.\arabic{table}}

	\section{Mutually dependent inputs modelled through copulas} \label{app:data_driven_input_model}
	
	In \subsecref{PCE} we illustrated PCE for an input $\vZ$ with independent components being uniformly distributed in $[0,1]$. $\vZ$ was obtained from the true input $\vX$ by assuming either that the latter had independent components as well, and thus defining $Z_i=F_i(X_i)$, or that a transformation $\T$ to perform this mapping was available. 
	
	This section recalls known results in probability theory allowing one to specify $\FX$ in terms of its marginal distributions and a dependence function called the \emph{copula} of $\vX$. We focus in particular on regular vine copulas (R-vines), for which the transformation $\T$ can be computed numerically. R-vines provide a flexible class of dependence models for $\fX$. This will prove beneficial to the ability of the PCE models to predict statistics of the output accurately, compared to simpler dependence models. However, and perhaps counter-intuitively, the pointwise error may not decrease accordingly (see \subsecref{error_types} for details), especially when $\T$ is highly non-linear. Examples on simulated data and a discussion are provided in \secref{synthetic_data}.

	\subsection{Copulas and Sklar's theorem} \label{subapp:Copulas_and_Sklars_theorem}
	
	A $\xdim$-copula is defined as a $\xdim$-variate joint $\cdf$ $C:[0,\:1]^{\xdim}\rightarrow[0,1]$
	with uniform marginals in the unit interval, that is,
	\[
	C(1,\ldots,1,u_{i},1,\ldots,1)=u_{i}\quad\forall u_{i}\in[0,1],\quad\forall i=1,\ldots,\xdim.
	\]
	
	Sklar's theorem \cite{Sklar1959} guarantees that any $\xdim$-variate joint $\cdf$ can be expressed in terms of $\xdim$ marginals and a copula, specified separately.
	
	\begin{thm*}[\textbf{Sklar}]
		For any $\xdim$-variate $\cdf$ $\FX$ with marginals $F_{1},\ldots,\:F_{\xdim}$, a $\xdim$-copula $\CX$ exists, such that 
		\begin{equation}
		\FX(\vx)=\CX\FXext.
		\label{eq:Sklar-F-C-relation}
		\end{equation}
		Besides, $\CX$ is unique on $\Ran(F_{1})\times\ldots\times\Ran(F_{\xdim})$, where $\Ran$ is the range operator. In particular, $\CX$ is unique 	on $[0,\,1]^{\xdim}$ if all $F_{i}$ are continuous, and it is given by
		\begin{equation}
		\CX\uext=\FX\FXinvext.
		\label{eq:Sklar-C-from-F}
		\end{equation}
		Conversely, for any $\xdim$-copula $C$ and any set of $\xdim$ univariate $\cdf$s $F_{i}$, $i=1,\ldots,\,\xdim$, the function 	$F:\,\mathbb{R}^\xdim \rightarrow [0,\,1]$ defined by
		\begin{equation}
		F \xext:=C \FXext
		\label{eq:Sklar-F-C-relation2}
		\end{equation}
		is a $\xdim$-variate $\cdf$ with marginals $F_{1},\ldots,F_{\xdim}$. 
	\end{thm*}
	
	Throughout this study it is assumed that $\FX$ has continuous marginals $F_{i}$. The relation \eqrefp{Sklar-F-C-relation2} allows one to model any multivariate $\cdf$ $F$ by modelling separately $\xdim$ univariate $\cdf$s $F_{i}$ and a copula function $C$. One first models the marginals $F_{i}$, then transforms each $X_{i}$ into a uniform random variable $U_{i}=F_{i}(X_{i})$ by the so-called probability integral transform (PIT) 
	\begin{equation}
	\TU: \vX \mapsto \vU = \left( F_1(X_1),\,\ldots,F_M(X_M) \right)^{\tr}.
	\label{eq:Uniform-transformation}
	\end{equation}
	Finally, the copula $C$ of $\vX$ is obtained as the joint $\cdf$ of $\vU$. The copula models the dependence properties of the random vector. For instance, mutual independence is achieved by using the independence copula
	\begin{equation}
	C(\vu) = \prod_{i=1}^{\xdim} u_i.
	\end{equation}
	\tabref{pair_copula_cdfs} provides a list of $19$ different parametric families of pair copulas implemented in the VineCopulaMatlab toolbox \cite{Kurz2014_CDVine}, which was also used in this study. Details on copula theory and on various copula families can be found in \cite{Nelsen2006} and in \cite{Joe2015}.
	
	Sklar's theorem can be re-stated in terms of probability densities. If $\vX$ admits joint $\pdf$ $\displaystyle{\fX(\vx):=\frac{\partial^\xdim \FX(\vx)}{\partial x_1\ldots \partial x_M}}$ and copula density $\displaystyle{\cX(\vu):=\frac{\partial^\xdim \CX(\vu)}{\partial u_1\ldots \partial u_M}}$, $\vu \in [0,1]^\xdim$, then
	\begin{equation}
	\fX(\vx)=c\FXext\cdot\prod_{i=1}^{\xdim}f_{i}(x_{i}).
	\label{eq:f-c-rel}
	\end{equation}
	
	Once all marginal $\pdf$s $f_i$ and the corresponding $\cdf$s $F_i$ have been determined (see \subsubsecref{marginals_inference}), each data point $\vx^{(j)} \in \dX$ is mapped onto the unit hypercube by the PIT \eqrefp{Uniform-transformation}, obtaining a transformed data set $\dU$ of \emph{pseudo-observations} of $\vU$. The copula of $\vX$ can then be inferred on $\dU$.

	\subsection{Vine copulas} \label{subapp:Vine_Copulas}
	
	In high dimension $\xdim$, specifying a $\xdim$-copula which properly describes all pairwise and higher-order input dependencies may be challenging. Multivariate extensions of pair-copula families (e.g. Gaussian or Archimedean copulas) are often inadequate when $\xdim$ is large. In \cite{Joe1996} and later in \cite{Bedford2002} and \cite{Aas2009}, an alternative construction by multiplication of $2$-copulas was introduced. Copula models built in this way are called \emph{vine copulas}. Here we briefly introduce the vine copula formalism, referring to the references for details. 
	
	Let $\vu_{\overline{i}}$ be the vector obtained from the vector $\vu$ by removing its $i$-th component, \ie, $\vu_{\overline{i}} = (u_{1},\ldots,u_{i-1},u_{i+1},\ldots,u_{\xdim})^{\tr}$. Similarly, let $\vu_{\overline{\{i,j\}}}$ be the vector obtained by removing the $i$-th and $j$-th component, and so on. For a general subset $\mathcal{\A}\subset\{1,\ldots,\,\xdim\}$, $\vu_{\notA}$ is defined analogously. Also, $\FcondA$ and $\fcondA$ indicate in the following the joint $\cdf$ and $\pdf$ of the random vector $\vX_{\notA}$ conditioned on $\vX_{\A}$. In the following, $\A=\{i_{1},\ldots,i_{k}\}$ and $\notA=\{j_{1},\ldots,\,j_{l}\}$ form a partition of $\{1,\ldots,\,\xdim\}$, \ie, $\A\cup\notA=\{1,\ldots,\,\xdim\}$ and $\A\cap\notA=\emptyset$.
	
	Using \eqrefp{f-c-rel}, $\fcondA$ can be expressed as
	\begin{equation}
	\begin{aligned}\fcondA(\vx_{\notA}|\vx_{\A})= & \ccondA(F_{j_{1}|\A}(x_{j_{1}}|\vx_{\A}),\,F_{j_{2}|\A}(x_{j_{2}}|\vx_{\A}),\ldots,\,F_{j_{l}|\A}(x_{j_{l}}|\vx_{\A}))\\
	& \qquad \times\,\prod_{j\in\notA}f_{j|\A}(x_j|\vx_{\A}),
	\end{aligned}
	\label{eq:sklar-cond-f-c-rel}
	\end{equation}
	where $\ccondA$ is an $l$-copula density -- that of the conditional random variables $(X_{j_{1}|\A},\,X_{j_{2}|\A},\ldots,\,X_{j_{l}|\A})^{\tr}$ -- and $f_{j|\A}$ is the conditional $\pdf$ of $X_{j}$ given $\vX_{\A}$, $j\in \notA$. Following \cite{Joe1996}, the univariate conditional distributions $F_{j|\A}$ can be further expressed in terms of any conditional pair copula $C_{ji|\A\backslash\{i\}}$ between $X_{j|\A\backslash\{i\}}$ and $X_{i|\A\backslash\{i\}}$, $i\in\A$:
	\begin{equation}
	F_{j|\A}(x_{j}|\vx_{\A})=\frac{\partial C_{ji|A\backslash\{i\}}(u_{j},\,u_{i})}{\partial u_{i}}\big\vert_{(F_{j|\A\backslash\{i\}}(x_{j}|\vx_{\A\backslash\{i\}}),\,F_{i|\A\backslash\{i\}}(x_{i}|\vx_{\A\backslash\{i\}}))}.
	\label{eq:Joe-condF-as-pairC}
	\end{equation}
	
	An analogous relation readily follows for conditional densities:
	\begin{equation}
	\begin{aligned}f_{j|\A}(x_{j}|\vx_{\A})= & \frac{\partial F_{j|\A}(x_{j}|\vx_{\A})}{\partial x_{j}}\\
	= & c_{ji|A\backslash\{i\}}(F_{j|\A\backslash\{i\}}(x_{j}|\vx_{\A\backslash\{i\}}),\,F_{i|\A\backslash\{i\}}(x_{i}|\vx_{\A\backslash\{i\}}))\\
	& \qquad \times\,f_{j|\A\backslash\{i\}}(x_{j}|\vx_{\A\backslash\{i\}}).
	\end{aligned}
	\label{eq:Joe-condf-as-pairc}
	\end{equation}
	
	By substituting iteratively \eqrefp{Joe-condF-as-pairC} and \eqrefp{Joe-condf-as-pairc}
	into \eqrefp{sklar-cond-f-c-rel}, \cite{Bedford2002} expressed $\fX$
	as a product of pair copula densities multiplied by $\prod_{i}f_{i}$.
	Recalling (\ref{eq:f-c-rel}), it readily follows that the associated
	joint copula density $c$ can be factorised into pair copula densities.
	Copulas expressed in this way are called vine copulas.
	
	The factorisation is not unique: the pair copulas involved in the construction depend on the variables chosen in the conditioning equations (\ref{eq:Joe-condF-as-pairC})-(\ref{eq:Joe-condf-as-pairc}) at each iteration. To organise them, \cite{Bedford2002} introduced a graphical model called the regular vine (R-vine). An R-vine among $\xdim$ random variables is represented by a graph consisting of $\xdim-1$ trees $T_{1},\,T_{2},\ldots,\,T_{\xdim-1}$, where each tree $T_{i}$ consists of a set $N_{i}$ of nodes and a set $E_{i}$ of edges $e=(j,k)$ between nodes $j$ and $k$. The trees $T_{i}$ satisfy the following three conditions:
	\begin{enumerate}
		\item Tree $T_{1}$ has nodes $N_{1}=\{1,\text{\ensuremath{\ldots},\,\xdim}\}$
		and $\xdim-1$ edges $E_{1}$
		\item for $i=2,\ldots,\,\xdim-1$, the nodes of $T_{i}$ are the edges of $T_{i-1}$
		: $N_{i}=E_{i-1}$
		\item Two edges in tree $T_{i}$ can be joined as nodes of tree $T_{i+1}$
		by an edge only if they share a common node in $T_{i}$ (proximity
		condition)
	\end{enumerate}
	To build an R-vine with nodes $\mathscr{N}=\{N_{1},\ldots,N_{\xdim-1}\}$
	and edges $\mathscr{E}=\{E_{1},\ldots,E_{\xdim-1}\}$, one defines for
	each edge $e$ linking nodes $j=j(e)$ and $k=k(e)$ in tree $T_{i}$,
	the sets $I(e)$ and $D(e)$ as follows:
	\begin{itemize}
		\item If $e\in E_{1}$ (edge of tree $T_{1}$), then $I(e)=\{j,\, k\}$ and $D(e)=\emptyset$,
		\item If $e\in E_{i}$, $i \geq 2$, then $D(e)=D(j) \cup D(k) \cup (I(j) \cap I(k))$ and $I(e) = (I(j) \cup I(k)) \backslash D(e)$.
	\end{itemize}
	$I(e)$ contains always two indices $j_e$ and $k_e$, while $D(e)$ contains $i-1$ indices for $e\in E_i$. One then associates each edge $e$ with the conditional pair copula $C_{j_e, k_e|D(e)}$ between $X_{j_e}$ and $X_{k_e}$ conditioned on the variables with indices in $D(e)$. An R-vine copula density with $\xdim$ nodes can thus be expressed as \cite{Aas2016}
	\begin{equation}
	c(\vu)=\prod_{i=1}^{\xdim-1} \prod_{e \in E_i} c_{j_e, k_e|D(e)}(u_{j_e|D(e)},\,u_{k_e|D(e)}).
	\label{eq:R-vine}
	\end{equation}
	
	Two special classes of R-vines are the drawable vine (D-vine; \cite{KurowickaCooke2005_inbook}) and the canonical vine (C-vine; \cite{Aas2009}). Denoting $F(x_{i})=u_{i}$ and $F_{i|\A}(x_{i}|\vx_{\A})=u_{i|\A}$, $i\notin\A$, a C-vine density is given by the expression 
	\begin{equation}
	c(\vu)=\prod_{j=1}^{\xdim-1}\prod_{i=1}^{\xdim-j}c_{j,j+i|\{1,\ldots,j-1\}}(u_{j|\{1,\ldots,j-1\}},\,u_{j+i|\{1,\ldots,j-1\}}),
	\label{eq:C-vine}
	\end{equation}
	while a D-vine density is expressed by
	\begin{equation}
	c(\vu)=\prod_{j=1}^{\xdim-1}\prod_{i=1}^{\xdim-j}c_{i,i+j|\{i+1,\ldots,i+j-1\}}(u_{i|\{i+1,\ldots,i+j-1\}},\,u_{i+j|\{i+1,\ldots,i+j-1\}}).
	\label{eq:D-vine}
	\end{equation}
	
	\begin{figure}
		\linespread{1}
		\begin{centering}
			\includegraphics[width=8cm]{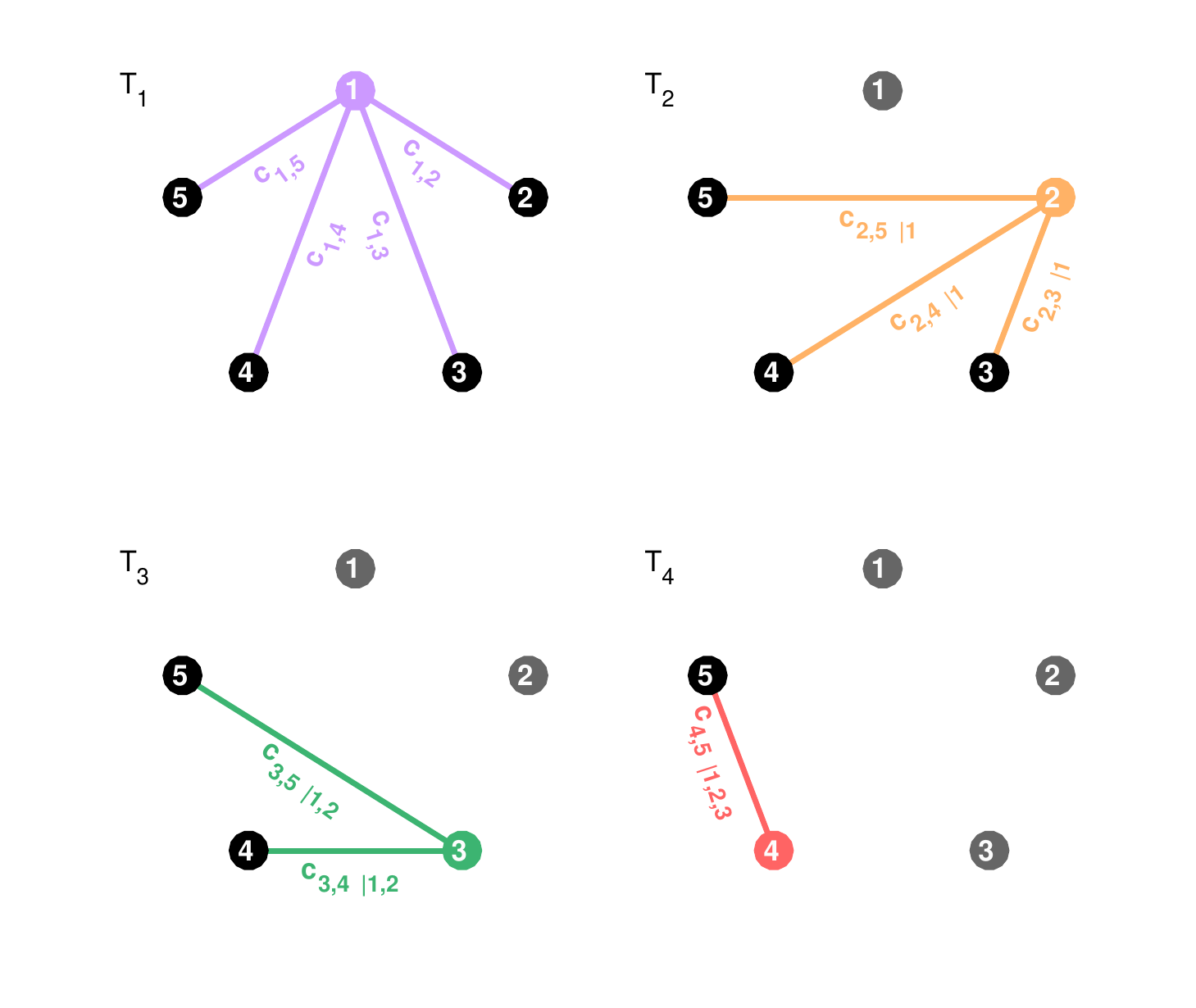}
			\includegraphics[width=4cm]{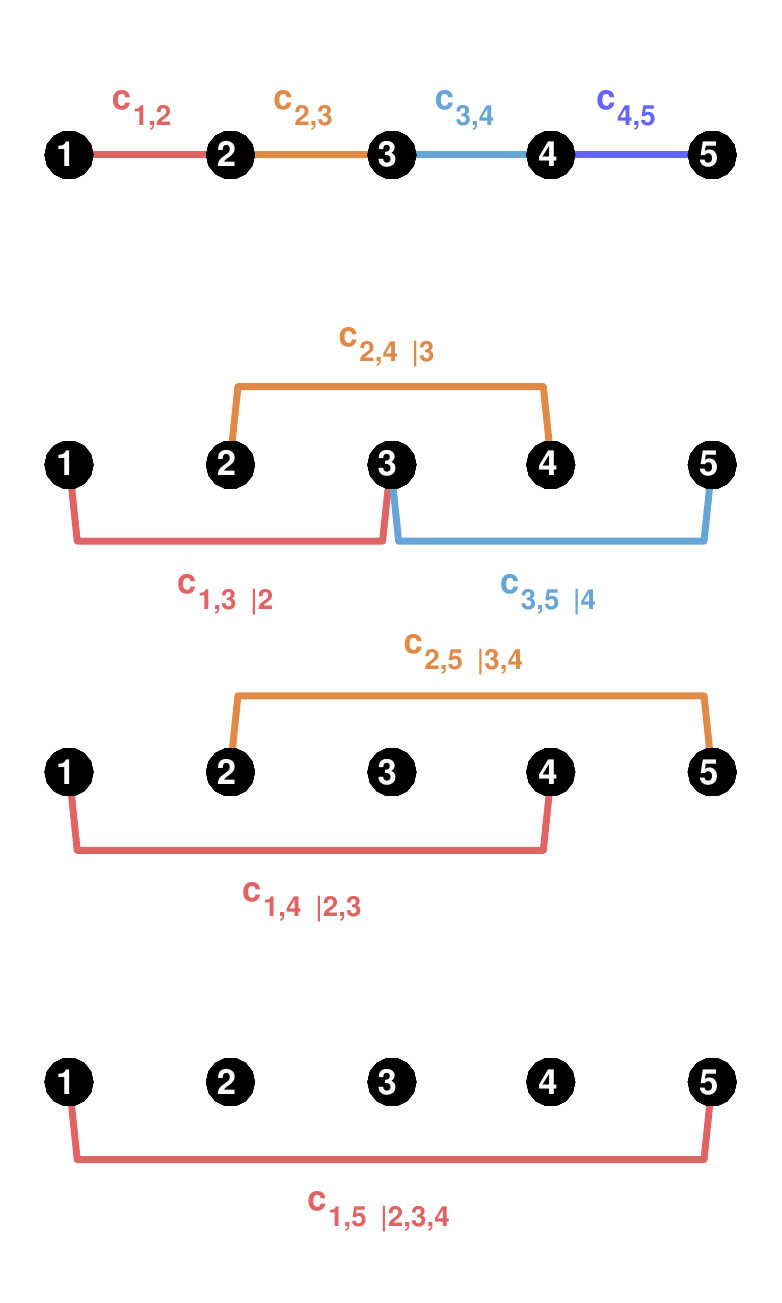}
			\caption{\textbf{Graphical representation of C- and D-vines.} The pair copulas
				in each tree of a $5$-dimensional C-vine (left; conditioning variables
				are shown in grey) and of a $5$-dimensional D-vine (right; conditioning
				variables are those between the connected nodes).}
			\label{fig:vines-5dim}
		\end{centering}
	\end{figure}
	
	The graphs associated to a $5$-dimensional C-vine and to a $5$-dimensional D-vine are shown in \figref{vines-5dim}. Note that this simplified illustration differs from the standard one introduced in \cite{Aas2009} and commonly used in the literature. 
	
	\subsection{Vine copula inference in practice}
	\label{subapp:Vine_construction_in_practice}
	
	We consider the purely data-driven case, typical in machine learning applications, where $\vX$ is only known through a set $\dX$ of independent observations. As remarked in \subsubsecref{marginals_inference}, inference on $\CX$ can be performed on $\dU$, obtained from $\dX$ by probability integral transform of each component after the marginals $f_i$ have been assigned. In this setup, building a vine copula model on $\dU$ involves the following steps:
	\begin{enumerate}
		\item Selecting a vine structure (for C- and D-vines: selecting the order of the nodes);
		\item Selecting the parametric family of each pair copula;
		\item Fitting the pair copula parameters to $\dU$.
	\end{enumerate}
	
	Steps 1-2 form the representation problem. Concerning step 3, algorithms to compute the likelihood of C- and D-vines \cite{Aas2009} and of R-vines \cite{Joe2015} given a data set $\dU$ exist, enabling parameter fitting based on maximum likelihood. In principle, the vine copula that best fits the data may be determined by iterating the maximum fitting approach over all possible vine structures and all possible parametric families of comprising pair copulas. In practice however, this approach is computationally infeasible in even moderate dimension $d$ due to the large number of possible structures \cite{MoralesNapoles2011_inbook} and of pair copulas comprising the vine. 
	
	Taking a different approach first suggested in \cite{Aas2009} and common to many applied studies, we first solve step $1$ separately. The optimal vine structure is selected heuristically so as to capture first the pairs $(X_{i},\,X_{j})$ with the strongest dependence (which then fall in the upper trees of the vine). The Kendall's tau \cite{Kendall1994} is selected as such a measure of dependence, defined by
	\begin{equation}
	\tau_{ij} =\mathbb{P}((X_{i}-\tilde{X}_{i})(X_{j}-\tilde{X}_{j})>0)-\mathbb{P}((X_{i}-\tilde{X}_{i})(X_{j}-\tilde{X}_{j})<0),
	\label{eq:def_Kendall_tau}
	\end{equation}
	where $(\tilde{X_{i}},\,\tilde{X_{j}})$ is an independent copy of $(X_{i},\,X_{j})$. If the copula of $(X_{i},\,X_{j})$ is $C_{ij}$, then 
	\begin{equation}
	\tau_{K}(X_i, X_j)=4\int\int_{[0,1]^{2}}C_{ij}(u,\,v)dC_{ij}(u,v)-1.
	\label{eq:Kendalls_tau_from_C}
	\end{equation}
	
	For a C-vine, ordering the variables $X_1,\,\ldots,X_M$ in decreasing order of dependence strength corresponds to select the central node in tree $T_{1}$ as the variable $X_{i_{1}}$ which maximises $\sum_{j\neq i_{1}}\tau_{i_{1}j}$, then the node of tree $T_{2}$ as the variable $X_{i_{2}}$ which maximises $\sum_{j\notin\{i_{1},i_{2}\}}\tau_{i_{2}j}$, and so on. For a D-vine, this means ordering the variables $X_{i_{1}},X_{i_{2}},\ldots,\,X_{i_{\xdim}}$ in the first tree so as to maximise $\sum_{k=1}^{\xdim-1}\tau_{i_{k}i_{k+1}}$, which we solve as an open travelling salesman problem  (OTSP) \cite{Applegate2006}. An open source Matlab implementation of a genetic algorithm to solve the OTSP is provided in \cite{Kirk2014_OTSP}. An algorithm to find the optimal structure for R-vines has been proposed in \cite{Dissmann2013_52}.
	
	For a selected vine structure, the vine copula with that structure that best fits the data is inferred by iterating, for each pair copula forming the vine, steps 2 and 3 over a list of pre-defined parametric families and their rotated versions. The families considered for inference in this paper are listed in \tabref{pair_copula_cdfs}. The rotations of a pair copula $C$ by $90$, $180$ and $270$ degrees are defined, respectively, by
	\begin{eqnarray}
	C^{(90)}(u,v) &=& v - C(1-u, v), \nonumber \\  
	C^{(180)}(u,v) &=& u+v-1 + C(1-u,1-v), \label{eq:rotated_copulas} \\
	C^{(270)}(u,v) &=& u - C(u, 1-v). \nonumber 
	\end{eqnarray}
	(Note that $C^{(90)}$ and $C^{(270)}$ are obtained by flipping the copula density $c$ around the horizontal and vertical axis, respectively; some references provide the formulas for actual rotations: ${C^{(90)}(u,v)=v-C(v, 1-u)}$, ${C^{(270)}(u,v)=u-C(1-v,u)}$). Including the rotated copulas, $62$ families were considered in total for inference in our study. 
	
	To facilitate inference, we rely on the so-called \emph{simplifying assumption} commonly adopted for vine copulas, namely that the pair copulas $C_{j(e),k(e)|D(e)}$ in \eqrefp{R-vine} only depend on the variables with indices in $D(e)$ through the arguments $F_{i(e)|D(e)}$ and $F_{j(e)|D(e)}$ \cite{Czado2010}. While being exact only in particular cases, this assumption is usually not severe \cite{Haff2010_simplifiedPCC}. Estimation techniques for non-simplified vine models have also been proposed \cite{Stoeber2013_101}.
	
	For each pair copula composing the vine, its parametric family is selected as the family that minimises the Akaike information criterion (AIC)
	\begin{equation}
	\AIC=2\times(k-\log L),
	\end{equation}
	where $k$ is the number of parameters of the pair copula and $\log L$ is its log-likelihood. 
	The process is iterated over all pair copulas forming the vine, starting from the unconditional copulas in the top tree (sequential inference). Finally (and optionally), one keeps the vine structure and the copula families obtained in this way (that is, the parametric form of the vine), and performs global likelihood maximisation.
	
	\begin{table}
		\linespread{1}
		\begin{center}
			\resizebox{\textwidth}{!}{
				\setlength{\extrarowheight}{1em}
				\begin{tabular}{llll}			
					\toprule
					ID & Name 
					& $\cdf$ 
					& Parameter range \\
					\cmidrule{1-4}
					1  & AMH
					& $\displaystyle{
						\frac{uv}{1-\theta(1-u)(1-v)}}$ 
					& $\theta \in [-1, 1]$ \\
					2  & AsymFGM 
					& $\displaystyle{uv \left( 1+\theta(1-u)^2 v (1-v) \right)}$ 
					& $\theta \in [0, 1]$ \\
					3  & BB1 
					& $\displaystyle{
						\left( 1+ \left( (u^{-\theta_2}-1)^{\theta_1} + (v^{-\theta_2}-1)^{\theta_1} \right)^{1/\theta_1} \right)^{-1/\theta_2}}$ 
					& $\theta_1 \geq 1$, $\theta_2 > 0$  \\
					4  & BB6 
					& $\displaystyle{
						1 - \left(1-\exp\left\lbrace -\left[ (-\log(1-(1-u)^{\theta_2}))^{\theta_1} + (-\log(1-(1-v)^{\theta_2}))^{\theta_1} \right]^{1/{\theta_1}} \right\rbrace\right)^{1/\theta_2}}$ 
					& $\theta_1 \geq 1$, $\theta_2 \geq 1$  \\
					5  & BB7 
					& $\displaystyle{
						\varphi(\varphi^{-1}(u)+\varphi^{-1}(v))}$, where 
					$\varphi(w)=\varphi(w; \theta_1,\theta_2)=1-\left(1-(1+w)^{-1/\theta_1}\right)^{1/\theta_2}$ 
					& $\theta_1 \geq 1$, $\theta_2 >0$ \\
					6  & BB8 
					& $\displaystyle{\frac{1}{\theta_1}\left( 1 - \left( 1 - \frac{(1-(1-\theta_1 u)^{\theta_2}) 
							(1-(1-\theta_1 v)^{\theta_2})}{1-(1-\theta_1)^{\theta_2}}\right)^{1/\theta_2} \right)}$ 
					& $\theta_1 \geq 1$, $\theta_2 \in (0, 1]$ \\  
					7  & Clayton 
					& $\displaystyle{
						(u^{-\theta} + v^{-\theta} -1)^{-1/\theta}}$ 
					& $\theta > 0$ \\
					8  & FGM 
					& $\displaystyle{uv(1+\theta(1-u)(1-v))}$  
					& $\theta \in (-1, 1)$ \\                     
					9  & Frank 
					& $\displaystyle{-\frac{1}{\theta} \log \left( \frac{1-e^{-\theta} - (1-e^{-\theta u})(1-e^{-\theta v})}{1-e^{-\theta}} \right)}	$ 
					& $\theta \in \mathbb{R} \backslash \lbrace 0 \rbrace$  \\
					10 & Gaussian 
					& $\displaystyle{\Phi_{2; \theta} \left(\Phi^{-1}(u), \Phi^{-1}(v)\right)}$ $^{(a)}$
					& $\theta \in (-1, 1)$  \\      
					11 & Gumbel
					& $\exp\left( - ((-\log u)^\theta + (-\log v)^\theta)^{1/\theta} \right) $ 
					& $\theta \in [1, +\inf)$ \\
					12 & Iterated FGM 
					& $uv(1+\theta_1(1-u)(1-v) + \theta_2 uv(1-u)(1-v))$ 
					& $\theta_1, \theta_2 \in (-1, 1)$ \\  
					13 & Joe/B5 
					& $1-\left( (1-u)^\theta + (1-v)^\theta +(1-u)^\theta (1-v)^\theta  \right)^{1/\theta}$ 
					& $\theta \geq 1$ \\
					14 & Partial Frank 
					& $\displaystyle{\frac{uv}{\theta(u+v-uv)} (\log(1+(e^{-\theta}-1)(1+uv-u-v)) + \theta)}$
					& $\theta > 0$  \\[.75em]
					15 & Plackett 
					& $\displaystyle{\frac{1+(\theta-1)(u+v)- \sqrt{(1+(\theta-1)(u+v))^2-4\theta(\theta-1)uv}}{2(\theta-1)}}$
					& $\theta \geq 0$ \\
					16 & Tawn-1 
					& $\displaystyle{(uv)^{A\left( \frac{\log v}{\log(uv)}; \theta_1,\theta_3 \right)}}$, where 
					$\displaystyle{A(w; \theta_1,\theta_3)=(1-\theta_3)w+\left[ w^{\theta_1} + (\theta_3(1-w))^{\theta_1} \right]^{1/\theta_1}}$
					& $\theta_1 \geq 1$, $\theta_3 \in [0,1]$ \\
					17 & Tawn-2 
					& $\displaystyle{(uv)^{A\left( \frac{\log v}{\log(uv)}; \theta_1,\theta_2 \right)}}$, where 
					$\displaystyle{A(w; \theta_1,\theta_2)=(1-\theta_2)(1-w)+\left[ (\theta_2 w)^{\theta_1} + ((1-w))^{\theta_1} \right]^{1/\theta_1}}$
					& $\theta_1 \geq 1$, $\theta_2 \in [0,1]$ \\
					18 & Tawn 
					& $\displaystyle{(uv)^{A\left( w; \theta_1,\theta_2,\theta_3 \right)}}$, where $\displaystyle{w=\frac{\log v}{\log(uv)}}$ and \\
					& & $\displaystyle{A(w; \theta_1,\theta_2,\theta_3)=(1-\theta_2)(1-w)+(1-\theta_3)w+\left[ (\theta_2 w)^{\theta_1} + (\theta_3(1-w))^{\theta_1} \right]^{1/\theta_1}}$
					& $\theta_1 \geq 1$, $\theta_2,\theta_3 \in [0,1]$ \\
					19 & t- 
					& $\displaystyle{t_{2;\nu,\theta}\left( t_{\nu}^{-1}(u), t_{\nu}^{-1}(v) \right)} $ $^{(b)}$
					& $\nu>1$, $\theta \in (-1,1)$ \\[.5em]
					\bottomrule
				\end{tabular}
			}
			\caption{\textbf{Distributions of bivariate copula families used in vine inference.} Copula IDs (reported as assigned in the VineCopulaMatlab toolbox), distributions, and parameter ranges. $(a)$ $\Phi$ is the univariate standard normal distribution, and $\Phi_{2;\theta}$ is the bivariate normal distribution with zero means, unit variances and correlation parameter $\theta$. $(b)$ $t_{\nu}$ is the univariate $t$ distribution with $\nu$ degrees of freedom, and $t_{\nu,\theta}$ is the bivariate $t$ distribution with $\nu$ degrees of freedom and correlation parameter $\theta$.}
			\label{tab:pair_copula_cdfs}
		\end{center}
	\end{table}

	\subsection{Rosenblatt transform of R-vines} \label{subapp:Rosenblatt_transforms}
	
	Suppose that the probabilistic input model $\FX$ is specified in terms of marginals $F_i$ and a copula $\CX$, the polynomial chaos representation can be more conveniently achieved by first mapping $\vX$ onto a vector $\vZ=\T(\vX)$ with independent components. 
	
	The most general map $\T$ of a random vector $\vX$ with dependent components onto a random vector $\vZ$ with mutually independent components is the Rosenblatt transform \cite{Rosenblatt52}
	\begin{equation}
	\TR: \, \vX \mapsto \bm{W},\,\textrm{where }\begin{cases}
	Z_{1}= & F_{1}(X_{1})\\
	Z_{2}= & F_{2|1}(X_{2}|X_{1})\\
	\;\vdots\\
	Z_{\xdim}= & F_{\xdim|1,\ldots,\xdim-1}(X_{\xdim}|X_{1},\ldots,\,X_{\xdim-1})
	\end{cases}.
	\label{eq:Rosenblatt-transform-1}
	\end{equation}
	
	One can rewrite (see also \cite{Lebrun2009c}) $\TR=\TPRa\circ\TU: \vX\mapsto\vU\mapsto\vZ$, where $\TU$, given by \eqrefp{Uniform-transformation}, is known once the marginals have been computed, while $\TPRa$ is given by
	\begin{equation}
	\TPRa: \, \vU \mapsto \vZ,\, \textrm{with } Z_i = C_{i|1,\ldots,i-1}(U_i|U_1,\ldots,U_{i-1}).
	\label{eq:Rosenblatt-transform-Copula-1}
	\end{equation}
	Here, $C_{i|1,\ldots,i-1}$ are conditional copulas of $\vX$ (and therefore of $\vU$), obtained from $\CX$ by differentiation. The variables $Z_{i}$ are mutually independent and have marginal uniform distributions in $[0,1]$. The problem of obtaining an isoprobabilistic transform of $\vX$ is hence reduced to the problem of computing derivatives of $\CX$.
	
	Representing $\CX$ as an R-vine solves this problem. Indeed, algorithms to compute \eqrefp{Rosenblatt-transform-Copula-1} have been established (see \cite{Schepsmeier2015}, and \cite{Aas2009} for algorithms specific to C- and D-vines). Given the pair-copulas $C_{ij}$ in the first tree of the vine, the algorithms first compute their derivatives $C_{i|j}$. Higher-order derivatives $C_{i|ijk}$, $C_{i|ijkh}$, $\ldots$ are obtained from the lower-order ones by inversion and differentiation. Derivatives and inverses of continuous pair copulas are generally computationally cheap to compute numerically, when analytical solutions are not available. The algorithms can be trivially implemented such that $n$ sample points are processed in parallel. 
	
	\setcounter{equation}{0}
	\section{PCE with R-vine input models} \label{app:vines_plus_PCE}
	
	If the inputs $\vX$ to the system are assumed to be mutually dependent, a possible approach to build a basis of orthogonal polynomials by tensor product is to transform $\vX$ into a random vector $\vZ$ with mutually independent components. The PCE metamodel can be built afterwards from $\vZ$ to $Y$. This approach, indicated by {\lPCEonZ} in the text, comprises the following steps: 
	\begin{enumerate}
		\item Model the joint $\cdf$ $\FX$ of the input by inferring its marginals and copula. Specifically:
		\begin{enumerate} \setlength{\itemsep}{0em}
			\item infer the marginal $\cdf$s $F_i$, $i=1,\ldots,\xdim$, from the input observations $\dX$ (\eg, by KDE);
			\item map $\dX$ onto $\dU=\{ \TU(\hat{\vx}^{(j)}),\, j=1,\ldots,n \} \subset [0,1]^d$ by \eqrefp{Uniform-transformation};
			\item model the copula $\CX \equiv C_{\vU}$ of the input by inference on $\dU$; R-vine copulas are compatible with this framework;
			\item define $\FX$ from the $F_i$ and $\CX$ using \eqref{eq:Sklar-F-C-relation}.
		\end{enumerate}
		\item Map $\dU$ onto $\dZ=\{ \TPRa(\hat{\vu}^{(j)}) \, j=1,\ldots,n \}$ by the Rosenblatt transform \eqref{eq:Rosenblatt-transform-Copula-1}. If the inferred marginals and copula are accurate, the underlying random vector $\vZ$ has approximately independent components, uniformly distributed in $[0,1]$.
		\item Build a PCE metamodel on the training set $(\dZ, \dY)$. Specifically, obtain the basis of $d$-variate orthogonal polynomials by tensor product of univariate ones. The procedure used to build each $i$-th univariate basis depends on the distribution assigned to $Z_i$ (if $Z_i \sim U([0,1])$, use Legendre polynomials; if $Z_i \sim \hat{F}_i$ obtained by KDE, use arbitrary PCE).
	\end{enumerate}
	
	The PCE metamodel obtained on the transformed input $\vZ=\T(\vX)$ can be seen as a transformation of $\vX$, 
	\begin{equation}
	Y_\PC (\vZ) = (Y_\PC \circ \T) (\vX). 
	\label{eq:YTPC}
	\end{equation}
	
\end{document}